\documentclass[10pt,journal,compsoc]{IEEEtran}
\usepackage{setspace}

\usepackage{url}

\usepackage{mathtools}
\usepackage{amsmath,amsopn,amssymb}
\usepackage{algorithm}  
\usepackage{algorithmicx}  
\usepackage{algpseudocode}  
\usepackage{amsmath}  

\usepackage{amsthm} 
\usepackage{verbatim}
\usepackage{balance}
\usepackage{amsfonts}
\usepackage{graphicx,xspace,color,soul}
\usepackage{epsfig,subfigure}
\usepackage{longtable,multirow}
\usepackage{array,float}
\usepackage{algorithmicx}
\usepackage{algorithm}
\usepackage{bm,epstopdf}
\usepackage{tabularx}
\usepackage{breqn}
\usepackage{hyperref}
\usepackage{changepage}

\usepackage{times}
\usepackage{epsfig}
\usepackage{amsfonts}       
\usepackage{booktabs}       
\usepackage{makecell}
\usepackage{algpseudocode}
\usepackage{pifont}

\usepackage{amsmath,amsfonts,bm}

\def\eqref#1{equation~\ref{#1}}

\def\1{\bm{1}}

\def\vf{{\bm{f}}}
\def\vg{{\bm{g}}}
\def\vh{{\bm{h}}}

\def\vn{{\bm{n}}}

\def\vs{{\bm{s}}}

\def\vw{{\bm{w}}}
\def\vx{{\bm{x}}}
\def\vy{{\bm{y}}}
\def\vz{{\bm{z}}}

\def\mF{{\bm{F}}}

\def\mH{{\bm{H}}}
\def\mI{{\bm{I}}}

\def\mL{{\bm{L}}}

\def\mN{{\bm{N}}}

\def\mX{{\bm{X}}}
\def\mY{{\bm{Y}}}
\def\mZ{{\bm{Z}}}

\DeclareMathAlphabet{\mathsfit}{\encodingdefault}{\sfdefault}{m}{sl}
\SetMathAlphabet{\mathsfit}{bold}{\encodingdefault}{\sfdefault}{bx}{n}

\def\gG{{\mathcal{G}}}
\def\gH{{\mathcal{H}}}

\def\gL{{\mathcal{L}}}
\def\gM{{\mathcal{M}}}
\def\gN{{\mathcal{N}}}

\def\gP{{\mathcal{P}}}

\def\gS{{\mathcal{S}}}

\def\gV{{\mathcal{V}}}

\def\sR{{\mathbb{R}}}

\newcommand{\E}{\mathbb{E}}

\newcolumntype{Y}{>{\centering\arraybackslash}X}

\def\ie{{\it i.e.}}
\def\eg{{\it e.g.}}
\def\et{{\it et al.}}
\def\etc{{\it etc.}}

\ifCLASSOPTIONcompsoc
  \usepackage[nocompress]{cite}
\else
  \usepackage{cite}
\fi

\begin{document}

\title{Deep Point Set Resampling via Gradient Fields}

\author{Haolan~Chen*,~\IEEEmembership{Student Member,~IEEE,}
Bi'an~Du*,~\IEEEmembership{Student Member,~IEEE,} \\
Shitong~Luo,~\IEEEmembership{Student Member,~IEEE,}
        and~Wei~Hu,~\IEEEmembership{Senior~Member,~IEEE}
        \thanks{* Equal contribution. H. Chen, B. Du, S. Luo and W. Hu are with Wangxuan Institute of Computer Technology, Peking University, No. 128, Zhongguancun North Street, Beijing, China (e-mail: \{chenhl99, scncdba, luost, forhuwei\}@pku.edu.cn).}
\thanks{Corresponding author: Wei Hu.}
}


\IEEEtitleabstractindextext{%
\begin{abstract}
3D point clouds acquired by scanning real-world objects or scenes have found a wide range of applications including immersive telepresence, autonomous driving, surveillance, etc. They are often perturbed by noise or suffer from low density, which obstructs downstream tasks such as surface reconstruction and understanding. In this paper, we propose a novel paradigm of point set resampling for restoration, which learns continuous gradient fields of point clouds that converge points towards the underlying surface. In particular, we represent a point cloud via its gradient field---the gradient of the log-probability density function, and enforce the gradient field to be continuous, thus guaranteeing the continuity of the model for solvable optimization. Based on the continuous gradient fields estimated via a proposed neural network, resampling a point cloud amounts to performing gradient-based Markov Chain Monte Carlo (MCMC) on the input noisy or sparse point cloud. Further, we propose to introduce regularization into the gradient-based MCMC during point cloud restoration, which essentially refines the intermediate resampled point cloud iteratively and accommodates various priors in the resampling process. Extensive experimental results demonstrate that the proposed point set resampling achieves the state-of-the-art performance in representative restoration tasks including point cloud denoising and upsampling.

\end{abstract}

}

\maketitle

\IEEEdisplaynontitleabstractindextext

%
\IEEEpeerreviewmaketitle

\section{Introduction}
\label{sec:introduction}

The maturity of depth sensing, laser scanning and image processing has enabled convenient acquisition of 3D point clouds from real-world scenes\footnote{Commercial products include Microsoft Kinect (2010-2014), Intel RealSense (2015-),
Velodyne LiDAR (2007-2020), LiDAR scanner of Apple iPad Pro (2020-), \etc.}, which consist of discrete 3D points irregularly sampled from continuous surfaces. 
Point clouds have attracted increasing attention as an effective representation of 3D shapes, which are widely applied in autonomous driving, robotics and immersive tele-presence. 
Nevertheless, they are often perturbed by noise or suffer from low density\footnote{This is evidenced in various public 3D scanning datasets such as KITTI \cite{geiger2013vision} and ScanNet \cite{dai2017scannet}.} due to the inherent limitations of scanning devices or matching ambiguities in the reconstruction from images, which significantly obstructs downstream tasks such as reconstruction and understanding.
Hence, point cloud restoration such as denoising and upsampling---essentially point set resampling---is crucial to relevant 3D vision applications.
However, point set resampling is challenging due to the irregular and unordered characteristics of point clouds.

Previous point cloud restoration works are mainly optimization based or deep-learning based. 
Optimization-based approaches rely heavily on geometric priors and are sometimes challenging to strike a balance between the detail preservation and restoration effectiveness, such as for point cloud denoising \cite{digne2017bilateral, huang2013bilat, cazals2005jetsfit, alexa2001MLS, avron2010sparsecoding, mattei2017MRPCA, sun2015lzero, zaman2017density} and upsampling \cite{alexa2003surfacefit,lipman2007parameterization,huang2009consolidation,huang2013bilat,wu2015modelnet}.    
Recently, deep-learning-based approaches have emerged and achieved promising restoration performance thanks to the advent of neural network architectures crafted for point clouds \cite{qi2017pointnet, qi2017pointnet2, wang2019dynamic}.
For point cloud denoising, the majority of deep-learning-based denoising models predict the displacement of noisy points from the underlying surface and then apply the inverse displacement to the noisy point clouds \cite{duan2019NeuralProj, rakotosaona2020PCN, hermosilla2019TotalDenoising, pistilli2020learning}.  
This class of methods mainly suffer from two types of artifacts: shrinkage and outliers, which arise from over-estimation or under-estimation of the displacement.  
Instead, Luo \et \cite{luo2021score} proposed score-based point cloud denoising, where the log-likelihood of each point is increased from the distribution of the noisy point cloud via gradient ascent----iteratively updating each point's position. 
Nevertheless, the gradient could be discontinuous that leads to unstable solutions, and no regularization has been introduced. 
For point cloud upsampling, complex regularization terms and fine-tuning are often required in order to prevent trivial upsampling results where points cluster together. 

\begin{figure}
\begin{center}
    \includegraphics[width=0.5\textwidth]{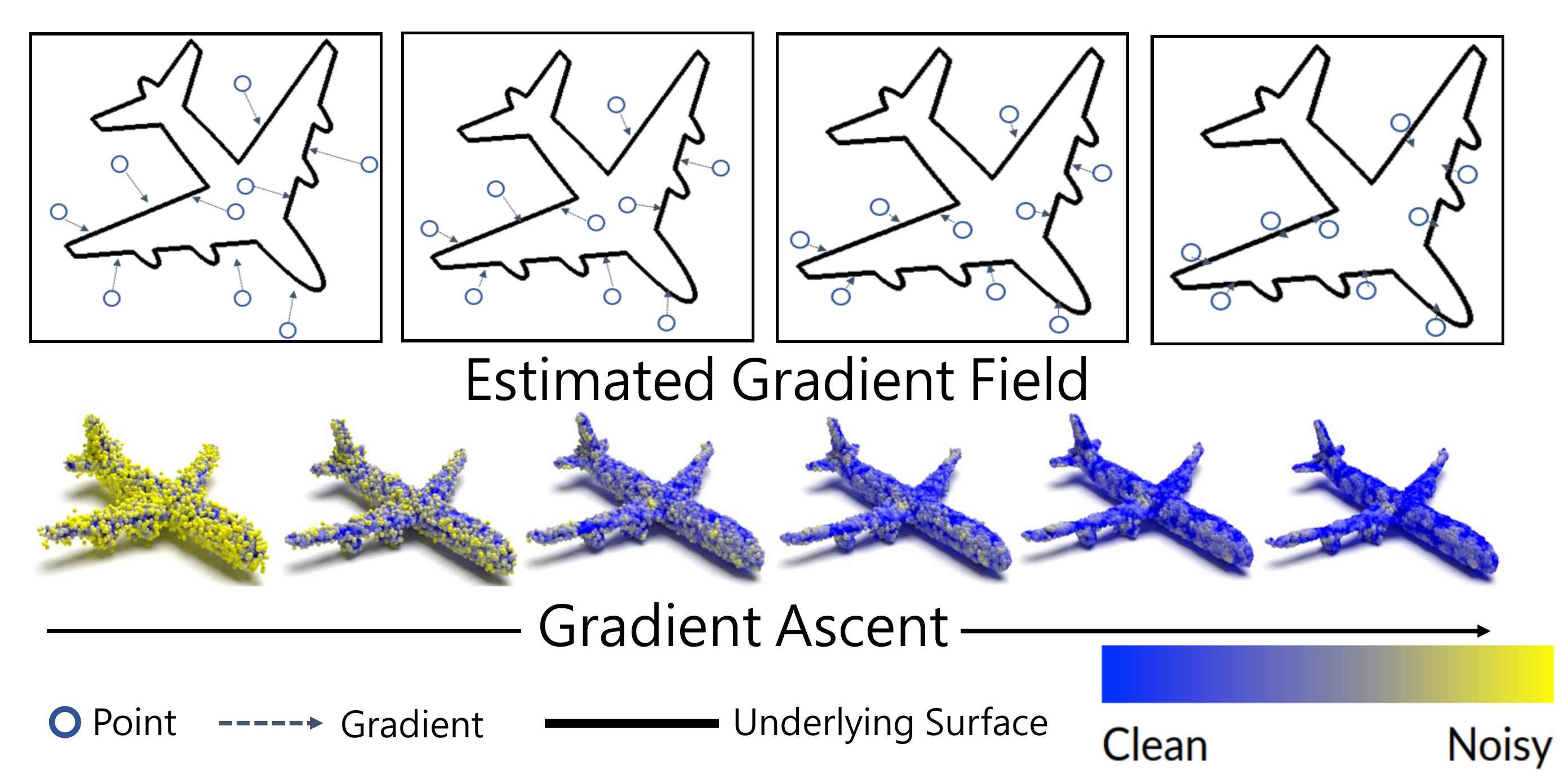}
\end{center}
\vspace{-0.25in}
   \caption{\textbf{An illustration of the proposed deep point set resampling method.} We first estimate the global gradient field of the degradation-convolved distribution $\nabla_\vx \log[ (p * h)(\vx) ]$ from the input degraded point cloud. Then, we perform gradient ascent using the estimated gradient field to converge points to the underlying surface for point cloud restoration. }
  \vspace{-0.25in}

\label{fig:teaser}
\end{figure}

To address these issues, we propose a novel paradigm of {\it deep point set resampling} for point cloud restoration, which models the distribution of degraded point clouds via gradient fields and converges points towards the underlying surface based on the learned gradient fields for restoration. 
Point clouds consist of discrete points $\vx$ sampled from the surface of 3D objects or scenes, and thus can be modeled as a set of samples from some 3D distribution $p(\vx)$ supported by 2D manifolds. 
If the point cloud is degraded from noise corruption and/or low density, the distribution of the degraded point cloud can be modeled as the convolution between the original distribution and some degradation model, expressed as $(p * h)(\vx)$.
The mode of $p * h$ is the underlying clean surface under some mild assumptions (see Section~\ref{subsec:model} for detailed analysis), having higher probability than its ambient space. 
According to this observation, restoring a degraded point cloud naturally amounts to moving perturbed points towards the mode, which can be realized by performing gradient ascent on the log-probability function $\log[(p * h)(\vx)]$, as illustrated in Figure~\ref{fig:teaser}. 
As the points are expected to converge to the mode of distribution after sufficient iterations of gradient ascent, our method is more robust against artifacts such as shrinkage and outliers, while previous methods have no awareness of the mode.

However, as $p * h$ is unknown at test-time, we instead estimate the global {\it gradient field} of the distribution underlying a degraded point cloud $\nabla_\vx \log[(p * h)(\vx)]$, \ie, the gradient of the log-probability density function.
Moreover, in order to guarantee solvable optimization via gradients, we enforce the gradient field to be {\it continuous} by cosine annealing. 
Further, unlike previous works where regularization by prior knowledge is incorporated into the loss function during the training stage, we propose to flexibly introduce regularization into the point set resampling process {\it subsequent to training}, which essentially refines the intermediate resampled point cloud iteratively during the sampling and is able to accommodate various priors such as the graph Laplacian regularizer (GLR) \cite{pang2017graph} for adaptive smoothing. 

Specifically, during the training stage, we take a context point cloud---the input degraded point cloud that provides the context of the underlying distribution, and develop a context feature extraction network and a gradient field estimation network to learn the gradient field of the distribution, which is optimized by our formulated objective function.
During the point cloud restoration stage, we resample the degraded point cloud by performing Markov Chain Monte Carlo (MCMC) such as Langevin dynamics based on the inferred gradient field. 
We also develop an algorithm to alternate the MCMC process and the regularization process iteratively until convergence. 

This work extends our previous work \cite{luo2021score} from the following four aspects. 
Firstly, instead of estimating a local gradient field around each point in the input as in \cite{luo2021score}, we propose to learn a {\it global} gradient field of the distribution supported over the entire point cloud.  
Secondly, we make the model {\it continuous} via cosine annealing, thus alleviating abrupt changes during the estimation of the global gradient field; 
Thirdly, we introduce regularization into the resampling process, thus enhancing the quality of the restored point clouds further; 
Finally, we generalize the application of point cloud denoising in \cite{luo2021score} to point cloud restoration such as denoising and upsampling, and perform extensive experiments under various noise types and upsampling ratios.

To summarize, our main contributions include
\begin{enumerate}
    \item We propose a novel paradigm of deep point set resampling for point cloud restoration, which models the distribution of degraded point clouds via global gradient fields and converges points towards the underlying surface for restoration.    
    \item We analyze the continuity of the distribution modeling and propose a continuous model by leveraging the cosine annealing, thus guaranteeing solvable optimization via gradients.    
    \item We introduce regularization into the point set resampling process, which is able to enhance the intermediate resampled point cloud iteratively during the sampling tailored for specific regularization designs flexibly.  
    
    \item Extensive experimental results demonstrate that the proposed point set resampling achieves the state-of-the-art performance in representative restoration tasks including point cloud denoising and upsampling, which also leads to satisfactory mesh reconstruction.    
\end{enumerate}

\section{Related Works}
\label{sec:related}
In this section, we discuss related works on point cloud restoration, including point cloud denoising and upsampling, respectively. 

\subsection{Point Cloud Denoising}
Point cloud denoising aims at restoring a clean point cloud from the noise-perturbed input, which can be classified into two categories: optimization-based methods and deep-learning-based methods.

\subsubsection{Optimization-based denoising}
This class of methods cast point cloud denoising as an optimization problem constrained by geometric priors. 
We classify them into four categories: 

(1) \textbf{Local-surface-fitting-based} methods approximate the point cloud with a smooth surface using simple-form function approximators explicitly or implicitly and then project points onto the surface \cite{alexa2001MLS}. 
Explicit methods include jet fitting \cite{cazals2005jetsfit} and bilateral filtering \cite{fleishman2003bilateral, huang2013bilat, digne2017bilateral} that take into account both point coordinates and normals. 
However, they are often sensitive to outliers and may fail to generate robust surface under extreme noise \cite{zheng2017guided}. 
Instead of parameterizing the underlying surface, implicit methods \cite{huang2009consolidation,lipman2007parameterization} generate a set of points that represent the underlying surface while preserving a uniform distribution. 
Nevertheless, they rely on local operators that tend to result in over-smoothing \cite{zheng2017guided}.

(2) \textbf{Sparsity-based} methods first reconstruct normals by solving an optimization problem with sparse regularization and then update the coordinates of points based on the reconstructed normals \cite{avron2010sparsecoding, sun2015lzero, xu2015sparsity}. The recently proposed MRPCA \cite{mattei2017MRPCA} is a sparsity-based denoiser which has achieved promising denoising performance.

(3) \textbf{Graph-based} methods abstract point clouds on graphs and perform denoising using graph filters such as the graph-Laplacian-based filters \cite{sch2015graphbased,zeng2019GLR, hu2020featuregraph, Hu2020gsp, hu2021dynamic}. Among them, Zeng \et \cite{zeng2019GLR} proposed graph Laplacian regularization (GLR) of a low-dimensional manifold model for point cloud denoising, while Hu \et \cite{hu2020featuregraph} proposed a paradigm of feature graph learning to infer the underlying graph structure of point clouds for denoising.

(4) \textbf{Density-based} methods are most relevant to ours as they also involve modeling the distribution of points. Zaman \et \cite{zaman2017density} deploys the kernel density estimation technique to approximate the density of noisy point clouds and focuses on outlier removal. Outlying points are removed in low-density regions. To finally obtain a clean point cloud, the bilateral filter \cite{fleishman2003bilateral} is leveraged to reduce the noise of the outlier-free point cloud. 

To summarize, optimization-based point cloud denoising methods rely heavily on geometric priors. Also, there is sometimes a trade-off between detail preservation and denoising effectiveness.

\subsubsection{Deep-learning-based denoising}
In recent years, neural network architectures tailored for irregular point cloud learning have emerged, \eg, PointNet \cite{qi2017pointnet}, PointNet++ \cite{qi2017pointnet2}, DGCNN \cite{wang2019dynamic}, \etc, which has made deep point cloud denoising possible. 
The majority of existing deep-learning-based methods predict the displacement of each point in noisy point clouds using neural networks, and apply the inverse displacement to each point. 
PointCleanNet (PCN) \cite{rakotosaona2020PCN} is the pioneer of this class of approaches, which employs a variant of PointNet as its backbone network \cite{qi2017pointnet} and minimizes the asymmetric Chamfer distance between the output and input point cloud.
GPDNet \cite{pistilli2020learning} uses graph convolutional networks to enhance the robustness of the neural denoiser.
NPD \cite{duan20193d} and PointProNets \cite{roveri2018PointProNets} also predict the position or the direction of local surfaces to guide the denoising process.
Hermosilla \et \cite{hermosilla2019TotalDenoising} proposed an unsupervised point cloud denoising framework---Total Denoising (TotalDn).
In TotalDn, an unsupervised loss function is derived for training deep-learning-based denoisers, based on the assumption that points with denser surroundings are closer to the underlying surface.
The aforementioned displacement-prediction methods generally suffer from two types of artifacts: shrinkage and outliers, as a result of inaccurate estimation of noise displacement.
Instead, Luo \et \cite{luo2020DMR} proposed to learn the underlying manifold (surface) of a noisy point cloud for reconstruction in a downsample-upsample architecture.
However, although the downsampling stage discards outliers in the input, it may also discard some informative details, leading to over-smoothing especially at low noise levels.
Motivated by the distribution model of noisy point clouds, Luo \et \cite{luo2021score} proposed to denoise point clouds via gradient ascent guided by the estimated gradient of the noisy point cloud's log-density, which distinguishes significantly from the aforementioned methods.
 
As discussed in the Introduction, we extend our previous work \cite{luo2021score} in four aspects. 
Our method is shown to alleviate the artifacts of shrinkage and outliers while preserving informative details, leading to significantly improved denoising performance.

\subsection{Point Cloud Upsampling}

Point cloud upsampling aims at generating dense point clouds from low-density data, which can also be classified into two categories: optimization-based methods and deep-learning-based methods.


\subsubsection{Optimization-based upsampling}
Prior to the emergence of deep-learning-based upsamplers, point cloud upsampling is often formulated as optimization problems with geometric prior constraints.
Alexa \et \cite{alexa2003surfacefit} proposed to upsample a point cloud by first constructing a Voronoi diagram, and then inserting new points at the vertices of the diagram.
Lipman \et \cite{lipman2007parameterization} proposed the locally optimal projection (LOP) operator to resample points and reconstruct surfaces based on an $L_1$ norm. 
Huang \et \cite{huang2009consolidation} proposed an improved weighted LOP and an iterative scheme to consolidate point clouds.
Huang \et \cite{huang2013bilat} further developed a progressive method called edge-aware resampling (EAR) for point cloud upsampling. 
However, EAR relies on the normals of the input point cloud, which requires extra estimation.
Wu \et \cite{wu2015deep} proposed a consolidation method but mainly focused on filling missing regions.
In summary, these methods rely heavily on appropriate geometric priors and require careful fine-tuning.

\subsubsection{Deep-learning-based upsampling}
The advent of point-based neural networks \cite{qi2017pointnet, qi2017pointnet2, wang2019dynamic} has also made deep point cloud upsampling possible, whose architectures lay the foundation for this task. 
Based on the PointNet++ architecture, Yu \et \cite{yu2018punet, yu2018ecnet} proposed PUNet and ECNet. PUNet \cite{yu2018punet} first encodes each point in the low-resolution point cloud into point-wise features. Then, the features are fed into a network to be expanded into new points. 
ECNet \cite{yu2018ecnet} improves PUNet by introducing a point-to-edge loss, aiming at preserving sharp features in the point cloud.
Wang \et \cite{wang2019MPU} proposed a cascaded upsampling network MPU, which consists of four similar sub-networks that are connected sequentially. 
A point cloud is upsampled by 2x after being passed through each of the sub-network.
More recently, Li \et \cite{li2019pugan} proposed PUGAN, which employs discriminators to fine-tune the upsampling network. 
Qian \et \cite{qian2020pugeo} designed PUGeo-Net that exploits the knowledge of differential geometry and learns to predict both coordinates and normals simultaneously.
Nevertheless, these models require complex regularization terms.



\section{Point Set Resampling Model}
\label{sec:model}


In this section, we will elaborate on the proposed modeling of point set resampling. 
We start from the basic mathematical modeling via gradient fields, and then analyze the continuity of the model. 
Finally, we introduce regularization into the model during the gradient-based resampling. 

\subsection{Distribution Modeling of Degraded Point Clouds}
\label{subsec:model}

To begin with, we view the distribution of an undegraded point cloud $\mY = \{\vy_i\}_{i=1}^N$ as sampled from a 3D distribution $p(\vy)$ supported by the underlying 2D manifolds. 
Since $\mY$ is discrete sampling from 2D manifolds, $p(\vy)$ is discontinuous and has zero support in the ambient space, \ie, $p(\vy) \rightarrow \infty$ if $\vy$ exactly lies on the manifold, otherwise $p(\vy) = 0$.

Next, we consider the distribution of {\it degraded} point clouds.
We denote a degraded point cloud as $\mX = \mH \circledast \mY + \mN$, where $\mH$ is a degradation function such as subsampling or blurring, and $\mN$ is an additive noise term from some noise distribution $\gN$ such as Gaussian distribution, while $\circledast$ denotes the convolution operation.
Here, we assume that the probability density function $\gN$ is continuous and has a unique mode at 0. These assumptions are made for analysis. 
We will show by experiments that in some cases where the assumptions do not hold, the proposed method still achieves superior performance (see Section~\ref{sec:experiments}). 
It can be shown that the clean point cloud $\mY$ from the distribution $p(\mY)$ exactly lies on the mode of $q(\mX)$ if the mode of $\gN$ is 0. 
When the assumption of uni-modality holds, $q(\mX)$ reaches the maximum on the manifold. 

Suppose the density function $q(\mX)$ is known. 
Based on the above analysis, restoring a point cloud $\mX = \{ \vx_i \}_{i=1}^N$ amounts to maximizing $\sum_i \log q(\vx_i)$.
This can be naturally achieved by performing gradient ascent until the points converge to the mode of $q(\vx)$. 
The gradient ascent relies \emph{only} on the {\it gradient field} $\nabla_\vx \log q(\vx)$---the first-order derivative of the log-density function.
As discussed above, $q(\vx)$ reaches the maximum on the underlying manifold under some mild assumptions.
Hence, the gradient field $\nabla_\vx \log q(\vx)$ consistently heads to the clean surface, as demonstrated in Figure~\ref{fig:teaser}. However, the density $q(\vx)$ is unknown during the test time. 
Instead of estimating $q(\vx)$ from degraded observations, we only resort to the gradient of $\log q(\vx)$, \ie, $\nabla_\vx \log q(\vx)$, which is more tractable.
This motivates the proposed model---deep point set resampling based on gradient fields.

Formally, our model aims at learning the gradient field $\vg(\vx)$ that leads to the maximization of $\sum_i \log q(\vx_i)$, \ie,
\begin{equation}
    \max_{\vg(\vx)} ~~~\sum_i \log q(\vx_i).
\end{equation}

It is obvious that points lying on the supporting manifolds satisfy $\vg(\vx)=0$. 
For points that are away from the supporting manifolds (\eg, due to noise perturbation), we estimate the gradients that converge points to the mode of the distribution $q(\vx)$, which corresponds to the underlying surface. 


\subsection{Continuity of the Model}
\label{sec:model_con}
As restoring the input point cloud is equivalent to solving the equation $\vg(\vx)=0$, continuity of the model is required to guarantee this equation can be solved iteratively via gradients.
Hence, we propose to employ cosine annealing that makes the estimation of the gradient field continuous with respect to the central points. 

In particular, as we estimate the gradient of some position $\vx$ from its local neighborhood $\gN_r(\vx)$ with radius $r$, when the position of $\vx$ changes during the resampling procedure, other points may enter or exit the neighborhood $\gN_r(\vx)$ abruptly, which will cause discontinuity. 
Hence, before aggregating features of nearby points, we assign each of them a corresponding weight, which decays as the distance from $\vx$ gets larger.
Formally, the aggregated feature of $\vx$ is 
\begin{equation}
\label{eq:vec_cos}
     \mF(\vx) = \sum_{\vx_j\in \gN_r(\vx)} \frac{1}{2} (\cos{ \pi\frac{\left|\vx - \vx_j\right|}{r} } + 1)\vf_{j}(\vx), 
\end{equation}
where $\vx_j \in \gN_r(\vx)$ denotes $\vx_j$ is adjacent to $\vx$ in the neighborhood $\gN_r(\vx)$, $\vf_{j}(\vx)$ is the relative feature of $\vx$ with respect to $\vx_j$, which we will estimate via a network as described later in Eq.~\ref{eq:gnet}. 
Essentially, the weight of the relative feature decays as the distance to $\vx$ enlarges, and finally becomes $0$ when the distance exceeds $r$.

Next, we prove that the above scheme makes the model continuous.
\begin{proof}
Assuming that when the coordinate of a point $\vx$ changes by $\Delta \vx$, and that there is only one point $\vx'$ that enters the neighborhood of point $\vx$, the change in the feature $\mF(\vx)$ is 
\begin{equation}
\label{eq:con}
\begin{split}
\lim\limits_{\Delta \vx \to 0 } \Delta \mF &= 
\lim\limits_{\Delta \vx \to 0 }  \frac{1}{2}  (\cos{ \pi\frac{\left|(\vx + \Delta \vx) - \vx' \right|}{r} + 1 )\vf_{j}(\vx)  }
\\
 &= \lim\limits_{\Delta \vx \to 0 } \frac{1}{2}  ( \cos{ \pi\frac{r - \delta(\Delta \vx)}{r}  } + 1)\vf_{j}(\vx) = 0,
\end{split}
\end{equation}
where $\delta$ is a very small distance related to $\Delta \vx$.
This indicates that the change in the feature converges to $0$ as the coordinate change of $\vx$ approaches $0$, which is continuous. 

If there are more than one point that enter or exit the neighborhood, we can similarly prove the above equation since the number of points must be limited. 
Hence, the continuity of our model is proved.
\end{proof}

\subsection{Regularization of the Model}


\begin{figure}
\begin{center}
    \includegraphics[width=0.5\textwidth]{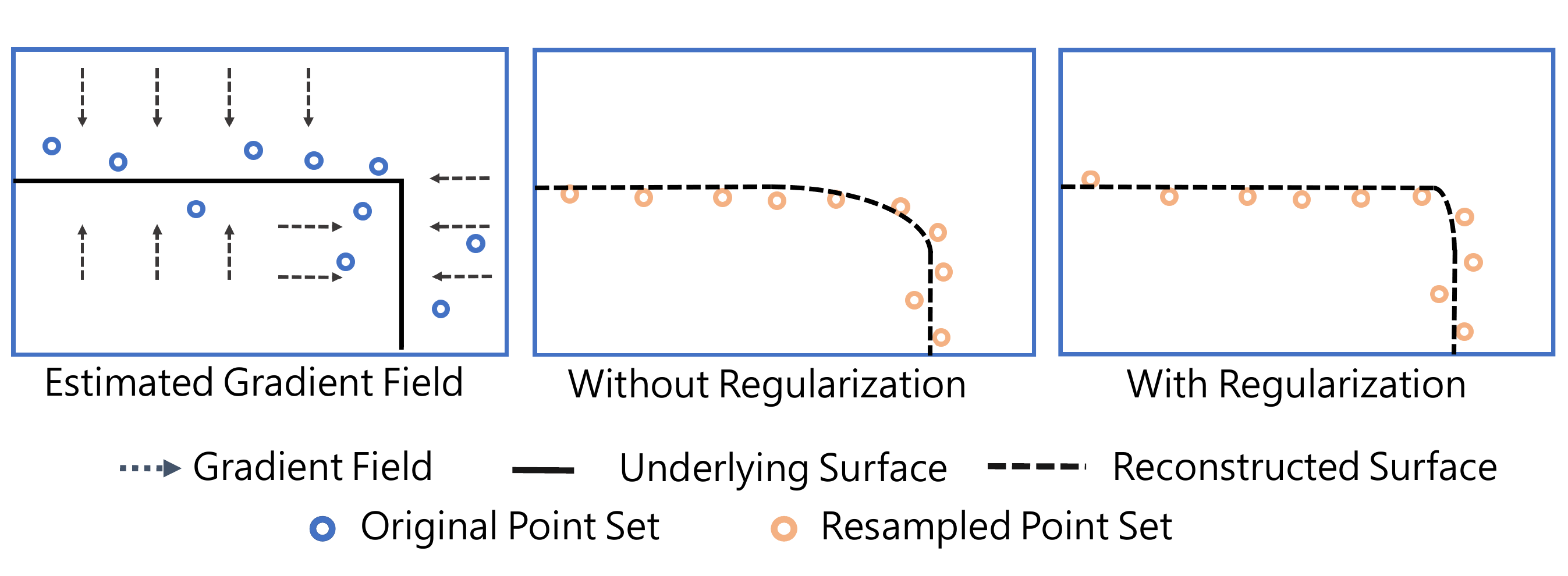}
\end{center}
\vspace{-0.2in}
\caption{{\bf A toy example to illustrate the comparison between resampling results without regularization and those with regularization.} 
An appropriate regularization leads to sharper boundaries.}
\label{fig:regularization}
\end{figure}

\label{subsec:regularization}
The proposed deep point set resampling framework first learns the gradient fields from data in the training stage, and then performs point cloud restoration via gradient ascent. 
This framework allows us to introduce regularization into the gradient ascent process for further refinement based on prior knowledge. 
Compared with existing works where regularization can only be considered in the training stage typically by incorporating into the loss function, our framework introduces regularization during the restoration process subsequent to training, which is thus more flexible for designing various priors for different downstream tasks. Figure~\ref{fig:regularization} illustrates a toy example where our resampling method preserves the sharpness of the surface better with the regularization considered.


\subsubsection{Formulation and Priors}
\label{subsubsec:prior}
Regularization enables restoring point clouds with desired properties from prior knowledge, such as piecewise smoothness \cite{hu2021graph}. 
A classical formulation to optimize a point cloud with prior knowledge is mathematically written as:
\begin{equation}
\label{eq:prior}
    \min_{\mZ} \;\| \mX - H(\mZ) \|_2^2 + \lambda \cdot \gP(\mZ),
\end{equation}
where $\mX$ and $\mZ$ denote the input degraded point cloud and the restored one, respectively. 
$H(\cdot)$ is a degradation operator (\eg, subsampling) defined over $\mZ$. 
$\gP(\mZ)$ represents a regularization term for $\mZ$.
$\lambda$ is a hyper-parameter to strike a balance between the first term of data fidelity and the second term of regularization.

There exist a variety of priors for point clouds, such as the Graph Laplacian Regularizer (GLR) \cite{pang2017graph} and the Reweighted Graph Laplacian Regularizer (RGLR) \cite{dinesh20183d} for preserving shape structures, and the repulsion prior \cite{wang2018repulsion} for enforcing points to be uniformly distributed, \etc. 
In this work, we focus on the GLR and RGLR, which are commonly adopted for point cloud restoration when abstracting point clouds on graphs in optimization-based approaches. 

In particular, graphs provide structure-adaptive, accurate, and compact representations for point clouds \cite{hu2021graph}. 
Hence, we represent each point in a point cloud as a node in a graph $\gG$, and connect nearby points to construct a graph, \eg, a $k$-nearest-neighbor ($k$NN) graph where each point is connected to its $k$ nearest neighbors. 
Then the coordinates of points serve as the graph signal over the graph. 
Specifically, given a graph signal $ \mZ $ residing on the vertices of a graph $ \mathcal{G} $ encoded in the graph Laplacian $\gL$ \cite{merris1994laplacian}, the GLR is mathematically expressed as:
\begin{equation}
\label{eq:glr}
    \gP_{\text{GLR}}(\mZ) = \mZ^{\top} \gL \mZ = \sum_{i\sim j}w_{i,j}  \| \vz_i - \vz_j \|_2^2,
\end{equation}
where $\gL$ is the graph Laplacian matrix that encodes the connectivity of the graph and the degree of each node. $i \sim j$ means vertices $i$ and $j$ are connected, implying the corresponding points on the geometry are highly correlated. $w_{i,j}$ is the weight of the edge connecting vertices $i$ and $j$. 
The signal $\mZ$ is {\it smooth} with respect to $\mathcal{G}$ if the GLR is small, as connected vertices $ \vz_i $ and $ \vz_j $ must be similar for a large edge weight $ w_{i,j} $ in order to minimize $\gP_{\text{GLR}}(\mZ)$ as in Eq.~\ref{eq:glr}; for a small $ w_{i,j} $, $\vz_i$ and $\vz_j$ can differ significantly. 

In the aforementioned GLR, the graph Laplacian $\gL$ is {\it fixed}, which does not promote restoration of the target signal with discontinuities if the corresponding edge weights are not very small. 
It is thus extended to \textit{Reweighted} GLR (RGLR) in \cite{liu2016random,pang2017graph,bai2018graph} by considering $\gL$ as a learnable function of the graph signal $\mZ$. 
The RGLR is defined as    
\begin{equation}
    \gP_{\text{RGLR}}(\mZ) = \mZ^{\top} \gL(\mZ) \mZ = \sum_{i\sim j}w_{i,j}(\vz_i,\vz_j)  \| \vz_i - \vz_j \|_2^2,
    \label{eq:rglr}
\end{equation}
where $w_{i,j}(\vx_i, \vx_j)$ can be learned from the data adaptively during the optimization process. 
Now there are two optimization variables $\mZ$ and $w_{i,j}$, which can be optimized alternately. 

It has been shown in \cite{bai2018graph} that minimizing the RGLR iteratively can promote piecewise smoothness in the reconstructed graph signal $\mZ$, assuming that the edge weights are appropriately initialized. 
Since point clouds often exhibit piecewise smoothness as discussed in \cite{hu2021graph}, the RGLR helps to promote this property in the restoration process.

\subsubsection{Optimization Solutions}
Having introduced various priors, we now discuss how to introduce the regularized optimization in Eq.~\ref{eq:prior} into the resampling process.
As most commonly used priors are differentiable (\eg, the GLR and RGLR), we focus on differentiable priors during the resampling. 

In Eq.~\ref{eq:prior}, assuming $H(\cdot)$ is differentiable, Eq.~\ref{eq:prior} exhibits a closed-form solution. 
For simplicity, we assume $H$ is an identity matrix (\eg, as in the denoising case). Then setting the derivative of Eq.~\ref{eq:prior} to zero yields
\begin{equation}
\label{eq:solution}
    2(\mX-\mZ) + \lambda \cdot \gP'(\mZ) = 0.
\end{equation}
Hence, $\mZ$ can be efficiently solved from Eq.~\ref{eq:solution}. 
Taking the GLR as an example, as $\gP_{\text{GLR}}'(\mZ) = 2\gL \mZ$, 
Eq.~\ref{eq:solution} admits the following closed-form solution:
\begin{equation}
\label{eq:closed_form_GLR}
    \mZ = (\mI+\lambda \cdot \gL)^{-1} \mX,
\end{equation}
where $\mI$ is an identity matrix. 
This is a set of linear equations and can be solved efficiently.
As $\gL$ is a high-pass operator \cite{hu2021graph}, the solution in Eq.~\ref{eq:closed_form_GLR} is essentially an adaptive low-pass filtering result from the observation $\mX$. 

We will present the proposed algorithm that alternates the regularization-based optimization and gradient ascent in Section~\ref{subsubsec:resample_regularization}.

\section{Point Set Resampling Algorithm} 
\label{sec:method}
\begin{figure}
\begin{center}
    \includegraphics[width=0.5\textwidth]{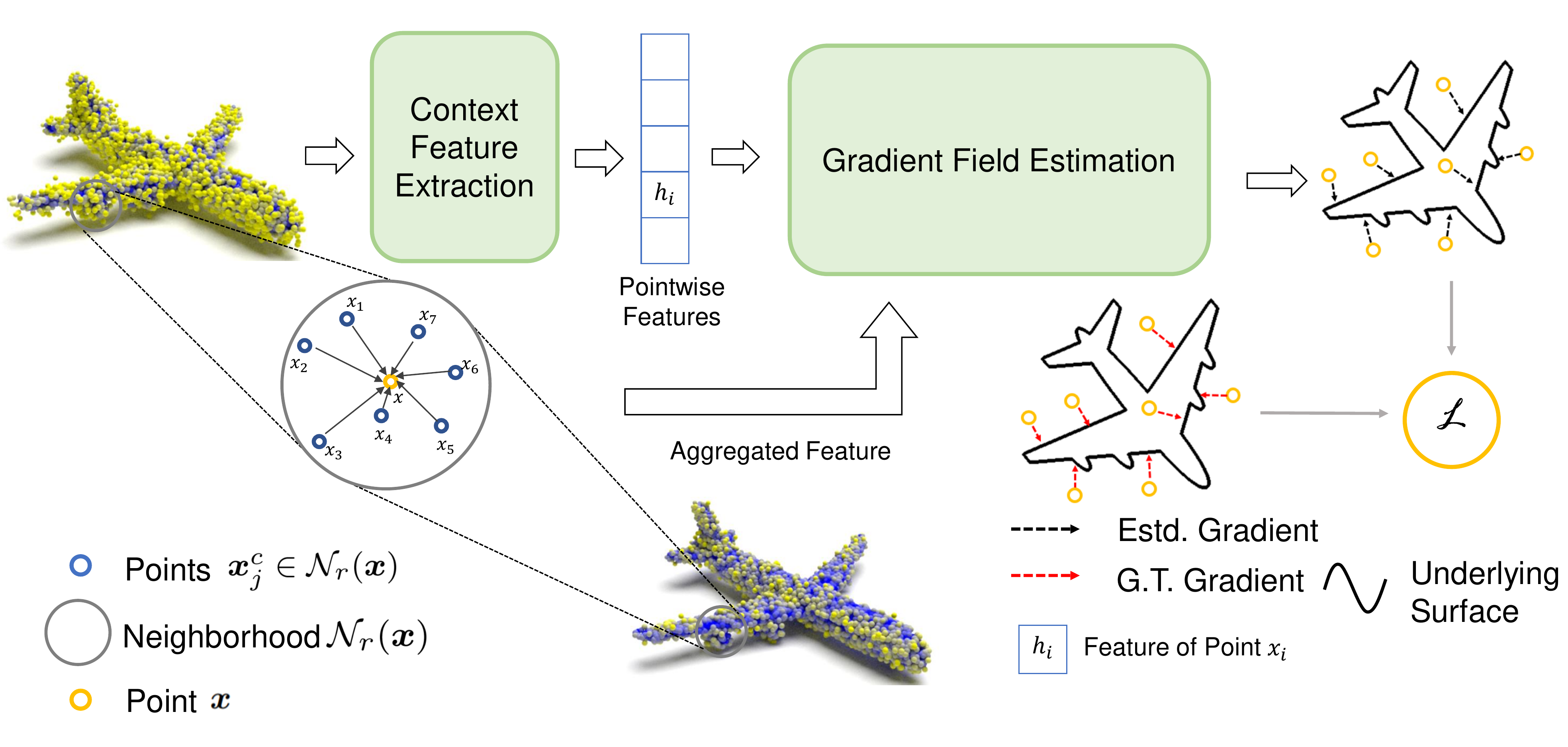}
\end{center}
\vspace{-0.25in}
  \caption{\textbf{An overview of the proposed network architecture.} Note that points $\vx_j^c\in \gN_r(\vx)$ are sampled from the context point cloud. }
\label{fig:overview}
\end{figure}

\begin{figure*}
\begin{center}
    \includegraphics[width=1.0\textwidth]{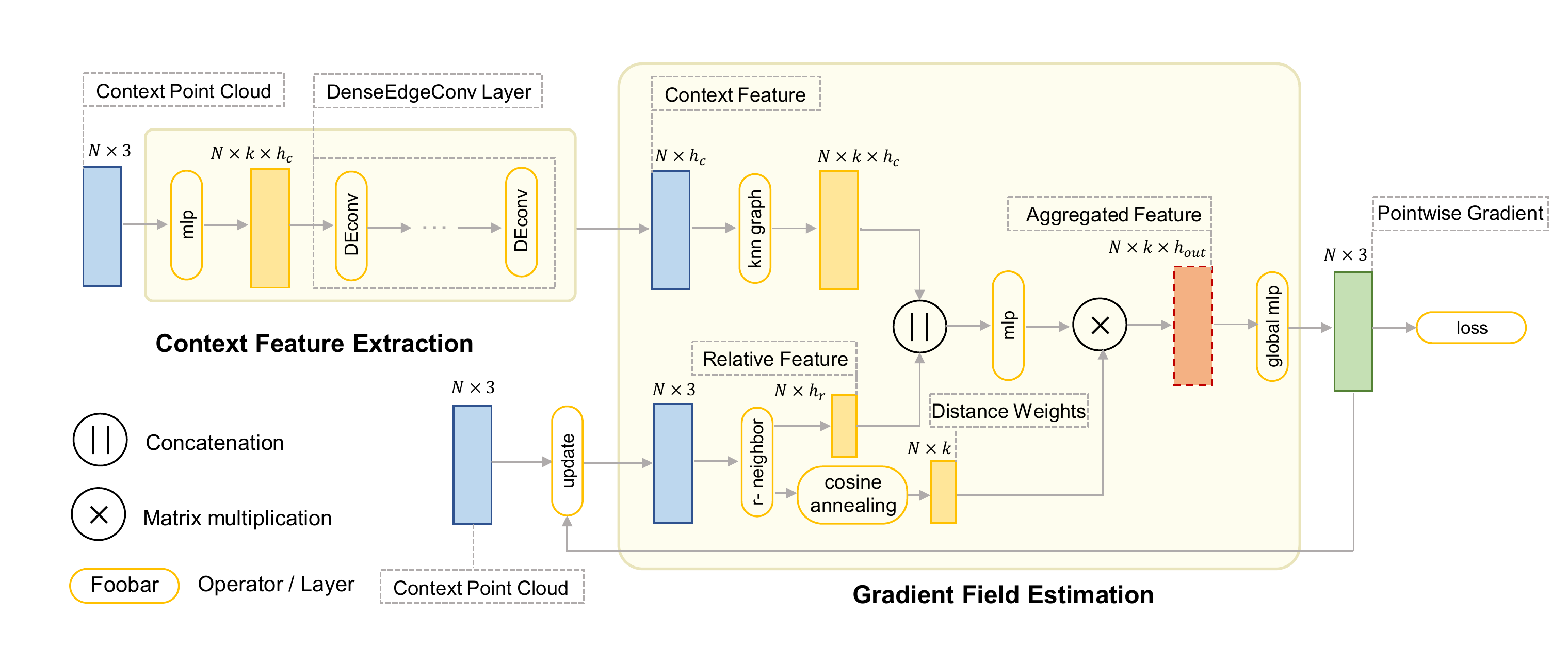}
\end{center}
\vspace{-0.2in}
\caption{{\bf An illustration of the proposed network architecture.} $h_{[*]}$ represents the feature dimension, where $h_c, h_r$ and $h_{out}$ represent the dimension of the context feature, relative feature and aggregated feature, respectively.}
\label{fig:architecture}
\end{figure*}
Based on the modeling of point set resampling in Section~\ref{sec:model}, we develop efficient algorithms for gradient field estimation and gradient-based resampling. 
Firstly, we will provide an overview of the entire architecture. 
Next, we discuss the network for the estimation of gradient fields and the training objective. 
Finally, we present the proposed resampling algorithm, without or with regularization introduced.  

\subsection{Overview}

Given a degraded point cloud $ \mX=\{\vx_i \}_{i=1}^{N}$ containing $N$ points, our goal is to perform point cloud restoration of the degraded $\mX$, such as denoising and upsampling, by point set resampling described in Sec~\ref{sec:model}.  
To implement the proposed deep point set resampling, we develop efficient networks for gradient field training and point cloud resampling respectively, which consist of the following modules as illustrated in Fig.~\ref{fig:overview}.
\begin{enumerate}
\setlength{\itemsep}{0pt}
\setlength{\parskip}{0pt}
    \item \textbf{Context Feature Extraction Network.} This module takes a context point cloud $\mX_C$ that provides the context of the underlying distribution as input and extracts point-wise features of $\mX_C$ for the subsequent gradient field estimation. 
    $\mX_C$ is assigned as the input degraded point cloud $\mX$ and kept fixed during the training (Section~\ref{subsubsec:feature}).
    
    \item \textbf{Gradient Field Estimation Network.} Based on the extracted point-wise features, this network estimates the gradient field underlying the degraded point cloud $\mX$ (Section~\ref{subsubsec:gradient}).
    
    \item \textbf{Training objective.} We design an objective for training the Gradient Field Estimation Network (Section~\ref{subs:obj}). 

    \item \textbf{Point Set Resampling.} We resample point sets by performing Markov Chain Monte Carlo (MCMC) based on the learned gradient field iteratively, with or without regularization (Section~\ref{subs:resample}).
\end{enumerate}

\subsection{The Proposed Training Network}
\label{subs:network}

In order to learn the gradient fields, our model first parameterizes the distribution of the supporting manifold underlying the context point cloud $\mX_C$ by feature extraction. 
The extracted feature is then employed to estimate the gradient fields pointing to the underlying surface. 
We discuss the context feature extraction network and the gradient field estimation network in order as follows.


\subsubsection{Context Feature Extraction Network}
\label{subsubsec:feature}
Given a context point cloud $\mX_C = \{\vx_i^c \}_{i=1}^{N}$, we first learn the features around each point $\vx_i^c$. 
In particular, we adopt the DGCNN \cite{wang2019dynamic} commonly used in previous denoising and upsampling models \cite{luo2020DMR, wang2019MPU, li2019pugan , luo2021score} as a basic unit, and build a stack of densely connected edge convolution layers for the context feature extraction.
Specifically, we construct a $k$NN graph over context point cloud $\mX_C$, where each point is treated as a vertex and connected to its $k$ nearest neighbors. 
Then we extract multi-scale as well as both local and non-local features for each point via the edge convolution layers, and further acquire features with richer contextual information via the dense connection \cite{huang2017densely, liu2019densepoint}.  
These properties make the architecture suitable for point cloud restoration tasks, as evidenced in previous works \cite{luo2020DMR, wang2019MPU,luo2021score}.
The learned feature for point $\vx_i^c$ is denoted as $\vh_i$. 

\subsubsection{Gradient Field Estimation Network}
\label{subsubsec:gradient}

Parameterized by point $\vx_i^c$'s feature $\vh_i$, this network aims to estimate a {\it global} gradient field of the distribution supported over the entire point cloud. Different from our previous work \cite{luo2021score} which estimates a gradient field {\it localized} around each point in the input context point cloud, we take some 3D coordinate $\vx \in \sR^3$ over the supporting manifold as input and output the global gradient $\vg(\vx)$, where $\vx$ does not necessarily correspond to a point in $\mX$.  

Specifically, we sample the $k$ nearest neighbors of some 3D coordinate $\vx$ in $\mX$, and employ the relative coordinates to learn relative features $\vf_{j}(\vx)$.
Formally, the gradient field estimation unit $\operatorname{F}$ takes the form:
\begin{equation}
\label{eq:gnet}
    \vf_{j}(\vx) = \operatorname{F}(\vx - \vx_j^c, \vh_j),
\end{equation}
where $\operatorname{F}(\cdot)$ is a multi-layer perceptron (MLP). 

To ensure the continuity of our model, a cosine annealing module that we introduced in Section~\ref{sec:model_con} (Eq.~\ref{eq:vec_cos}) is appended to aggregate the relative feature corresponding to each neighbor $\vx_j^c$ into the point-wise feature of $\vx$ with distance-related weights, leading to the aggregated feature $\mF(\vx)$. 

Finally, $\mF(\vx)$ is fed into a global MLP consisting of convolution layers to output the final gradient field. 
Then the final estimation of the gradient field of is  
\begin{equation}
\label{eq:vec}
    \vg(\vx) = \gM(\sum_{\vx_j^c\in \gN_r(\vx)} \frac{1}{2} (\cos{ \pi\frac{\left|\vx - \vx_j^c\right|}{r} } + 1) \operatorname{F}(\vx - \vx_j^c, \vh_i) ),
\end{equation}
where $\gM$ represents the global MLP.
The gradient field estimation is trained by optimizing the proposed objective, which will be discussed next.

\algrenewcommand\algorithmicindent{1.0em}%
\begin{algorithm}[t]
    \caption{Training of Gradient Field Learning} 
    \label{alg:train}
    \small
     \hspace*{0.02in} {\bf Input:}
        The degraded point cloud $\mX= \{\vx_i\}_{i=1}^N$, context point cloud $\mX_C= \{\vx_i^c\}_{i=1}^N$ set as $\mX_C=\mX$, and ground truth point cloud $\mY $. \\
     \hspace*{0.02in} {{\bf Initialize:}
        The Context Feature Extraction Module $\gH$; 
        the Gradient Field Estimation Unit $\operatorname{F}$;
        the global MLP $\gM$.}
    \begin{algorithmic}[1]
    \While {not converge}
    \State Construct a $k$NN graph $\gG$ over context point cloud $\mX_C$.
    \State Learn the point-wise context feature $\vh_i = \gH(\vx_i^c, \gG)$.
    \State Learn relative feature $\vf_{j}(\vx) = \operatorname{F}(\vx - \vx_j^c, \vh_j)$,$\vx_j^c\in \gN_r(\vx)$\
    \State Compute distance weight $\vw_{j}(\vx) = \frac{1}{2}(\cos{ \pi\frac{\left|\vx - \vx_j^c \right|}{r} } + 1)$.
    \State Aggregate relative feature $\mF(\vx) = \sum_{\vx_j ^c \in \gN_r(\vx)}\vw_{j}(\vx)\vf_{j}(\vx) $. 
    \State Predict the gradient field $\vg(\vx) = \gM(\mF(\vx))$ via the global MLP $\gM$.
    \State Compute the loss function in Eq.~\ref{eq:supervised}.
    \State Update network parameters with the SGD optimizer and $\mathcal{L}$.
    \EndWhile 
    \end{algorithmic}
    \hspace*{0.02in} {\bf Output:}
        $\gH$, $\operatorname{F}$, $\gM$ with trained weights. \\
    
\end{algorithm}
\subsection{The Training Objective}
\label{subs:obj}
We denote the input degraded point cloud as $\mX = \{ \vx_i\}_{i=1}^N$ and the ground truth point cloud\footnote{For example, in the point cloud denoising task, the ground truth is the clean point cloud.} as $\mY = \{ \vy_i \}_{i=1}^N$. 
Using the ground truth $\mY$, we define the gradient for some point $\vx \in \sR^3$ as follows:
\begin{equation}
\label{eq:gtscore}
    \vs(\vx) = \operatorname{NN}(\vx, \mY) - \vx, \ \vx \in \sR^3,
\end{equation}
where $\operatorname{NN}(\vx, \mY)$ returns the point nearest to $\vx$ in $\mY$.
Intuitively, $\vs(\vx)$ is a vector from $\vx$ to the underlying surface.

The training objective aligns the network-predicted gradient field to the ground truth defined above:
\begin{equation}
\label{eq:supervised}
    \mathcal{L} = \E_{\vx \sim \mathcal S} \left[ \left\| \vs(\vx) - \vg(\vx)  \right\|_2^2 \right],
\end{equation}
where $\gS$ is a distribution of $\vx$ in $\sR^3$ space.
Note that, this objective not only matches the predicted gradient field on the position of $\vx$ but also matches the gradient field on the neighboring areas of $\vx$.
This is important because a point moves around during gradient ascent for resampling, which relies on the gradient field defined over its neighborhood provided by the context point cloud $\mX_C$.
Such definition of training objective also distinguishes our method from previous displacement-based methods \cite{rakotosaona2020PCN, pistilli2020learning}, as the objectives of those methods only consider the position of each point while our objective covers the neighborhood of each point.

\subsection{Point Set Resampling}
\label{subs:resample}
In the point cloud restoration stage, given a degraded point cloud $\mX = \{ \vx_i \}_{i=1}^N$ as input, we first construct the gradient field $\vg(\vx)$ for the point cloud $\mX$.
Specifically, we first feed $\mX$ into the context feature extraction network to obtain a set of point-wise features $\{ \vh_i \}_{i=1}^N$.
Next, by substituting $\vx_i$, $\vh_i$ and some 3D coordinate $\vx \in \sR^3$ into Eq.~\ref{eq:vec}, we acquire $\vg(\vx)$ as the estimated gradient field.


Next, we perform point set resampling via gradient ascent to achieve point cloud restoration.
As discussed in Section~\ref{subsec:regularization}, our model opts to introduce regularization or not during resampling, depending on the requirement of specific tasks. 
Thus, we develop algorithms for resampling without regularization and with regularization respectively, which are presented as follows. 

\subsubsection{Resampling without regularization}

In the simple setting without regularization, restoring a point cloud amounts to updating each point's position via gradient ascent:
\begin{equation}
\label{eq:denoise}
\begin{split}
    \vx_i^{(0)} & = \vx_i, \ \vx_i \in \mX , \\
    \vx_i^{(t)} & = \vx_i^{(t-1)} + \alpha_t \vg(\vx_i^{(t-1)}), \ t = 1,\ldots,T , 
    \end{split}
\end{equation}
where $\alpha_t$ is the step size at the $t$-th step. 
We suggest two criteria for choosing the step size sequence $\{ \alpha_t \}_{t=1}^T$: 
(1) The sequence should be decreasing towards 0 to ensure convergence. 
(2) $\alpha_1$ should be less than 1 and not be too close to 0. This is because according to Eq.~\ref{eq:gtscore}, the magnitude of the score is approximately the distance from each point to the underlying surface (approximately the length of $\vs(\vx)$ in Eq.~\ref{eq:gtscore}). Thus, performing gradient ascent for a sufficient number of steps with a proper step size less than 1 is enough. The final reconstructed point cloud is $\hat{\mY} =\{\vx_i^{(T)}\}_{i=1}^N$.

\subsubsection{Resampling with regularization}
\label{subsubsec:resample_regularization}
\algrenewcommand\algorithmicindent{1.0em}%
\begin{algorithm}[t]
    \caption{Point Cloud Restoration} 
    \label{alg:restoration}
    \small
     \hspace*{0.02in} {\bf Input:}
        The degraded point cloud $\mX= \{\vx_i\}_{i=1}^N$, context point cloud $\mX_C= \{\vx_i^c\}_{i=1}^N$ set as $\mX_C=\mX$, number of steps $T$, step decay rate $d$; $\gH$, $\operatorname{F}$, $\gM$ with trained weights from {Algorithm 1}.\\
    \hspace*{0.02in} {\bf Initialize: }step size $\alpha_1$
    \begin{algorithmic}[1]    
    \State Extract context feature $\vh_i \gets \gH(\vx_i^c) $ 
    \State $t \gets 1 , \vx_i^{(0)} \gets \vx_i, \ \vx_i \in \mX $
    \While {$t < T + 1$}
    \State $\vf_{j}(\vx^{(t-1)}) \gets \operatorname{F}(\vx^{(t-1)} - \vx_j^c, \vh_j)$ , $\vx_j^c \in \gN_r(\vx^{(t-1)})$
    \State $\mF(\vx^{(t-1)}) \gets \sum_{\vx_j^{(t-1)}\in \gN_r(\vx^{(t-1)})}\vw_{j}(\vx^{(t-1)})\vf_{j}(\vx^{(t-1)}) $.
    \State Predict the gradient field $\vg(\vx^{(t-1)}) \gets \gM(\mF(\vx^{(t-1)}))$
    \If {Resampling with regularization} 
    \State $\tilde{\vx}_i^{(t)} \gets
    \vx_i^{(t-1)} + \alpha_{t} \vg(\vx_i^{(t-1)})$
    \State $\vx_i^{(t)} \gets
    (\mI+\lambda \cdot \gL)^{-1}\tilde{\vx}_i^{(t)}$
    \State $\alpha_{t+1} \gets d \alpha_t $
    \Else
    \State $\vx_i^{(t)} \gets \vx_i^{(t-1)} + \alpha_{t} \vg(\vx_i^{(t-1)})$
    \State $\alpha_{t+1} \gets d \alpha_t $
    \EndIf
    
    \EndWhile
    \State $\hat{\mY} \gets \{\vx_i^{(T)}\}_{i=1}^N$
    \end{algorithmic}
    \hspace*{0.02in} {\bf Output:}
        The restored point cloud $\hat{\mY}$.\\
\end{algorithm}
As regularization is often beneficial to point cloud restoration as discussed in Section~\ref{subsec:regularization}, we also develop a resampling algorithm with regularization considered. 
As mentioned, we focus on the commonly adopted regularizers for point clouds---the GLR and RGLR, which are introduced into the resampling process by alternating the gradient ascent and regularization-based optimization to exploit prior knowledge for the refinement of the intermediate resampled point cloud.

\noindent \textbf{Resampling with the GLR.}
As the GLR-based optimization admits a closed-form solution as in Eq.~\ref{eq:closed_form_GLR}, we alternate the gradient ascent and GLR-based optimization as follows:  
\begin{equation}
\label{eq:restoration_GLR}
\begin{split}
    \vx_i^{(0)} & = \vx_i, \ \vx_i \in \mX , \\
    \tilde{\vx}_i^{(t)} & = \vx_i^{(t-1)} + \alpha_t \vg(\vx_i^{(t-1)}), \\
    \vx_i^{(t)} & = (\mI+\lambda \cdot \gL)^{-1}\tilde{\vx}_i^{(t)} ,\ t = 1,\ldots,T ,
\end{split}
\end{equation}
where $\gL$ is computed from the input point cloud $\mX$ and kept fixed during the iterations. 

To compute $\gL$, we first construct a $k$NN graph over $\mX$. The edge weight $w_{i,j}$ connecting points $i$ and $j$ is assigned as a function of the Euclidean distance between $\vx_i$ and $\vx_j$:
\begin{equation}
\label{eq:weight}
    w_{i,j} = \exp \left\{-\frac{\|\vx_i-\vx_j\|_2^2}{\sigma^2} \right\},
\end{equation}
where $\sigma$ is a parameter.

Based on the iterative optimization in Eq.~\ref{eq:restoration_GLR}, the final reconstructed point cloud is $\hat{\mY} =\{\vx_i^{(T)}\}_{i=1}^N$.

\noindent \textbf{Resampling with the RGLR.}
In the RGLR, the graph Laplacian is dynamically updated from each intermediate resampled point cloud. 
As there exist two optimization variables in the RGLR-based optimization as presented in Eq.~\ref{eq:rglr}, there is no closed-form solution and alternate optimization is often adopted. 
Hence, we perform the following iterative process for resampling with the RGLR:
\begin{equation}
\label{eq:restoration_RGLR}
\begin{split}
    \vx_i^{(0)} & = \vx_i, \ \vx_i \in \mX , \\
    \tilde{\vx}_i^{(t)} & = \vx_i^{(t-1)} + \alpha_t \vg(\vx_i^{(t-1)}), \\
    \gL^{(t)} & \leftarrow \tilde{\mX}^{(t)}, \\
    \vx_i^{(t)} & = (\mI+\lambda \cdot \gL^{(t)})^{-1} \tilde{\vx}_i^{(t)} ,\ t = 1,\ldots,T ,
\end{split}
\end{equation}
where $\gL^{(t)}$ is updated from the optimized intermediate point cloud $\mX^{(t)}$. 
The initialization of $\gL^{(0)}$ is the same as in Eq.~\ref{eq:weight}, and the update of $\gL^{(t)}$ follows the same function of edge weights computed from $\mX^{(t)}$. 
The final reconstructed point cloud is $\hat{\mY} =\{\vx_i^{(T)}\}_{i=1}^N$.

Finally, we provide a summary of the training algorithm of gradient field estimation in Algorithm~\ref{alg:train} and the point cloud restoration algorithm in Algorithm~\ref{alg:restoration}. 
For the sake of simplicity, we present resampling with the GLR as a representative regularization in Algorithm~\ref{alg:restoration}.  

\begin{table*}
\begin{center}
\resizebox{0.99\textwidth}{!}{
\begin{tabular}{c|l| cccccc|cccccc}
\toprule
\multicolumn{2}{c|}{\# Points} & \multicolumn{6}{c|}{10K (Sparse)} & \multicolumn{6}{c}{50K (Dense)} \\

\multicolumn{2}{c|}{Noise} & 
\multicolumn{2}{c}{1\%} & \multicolumn{2}{c}{2\%} & \multicolumn{2}{c|}{3\%} &
\multicolumn{2}{c}{1\%} & \multicolumn{2}{c}{2\%} & \multicolumn{2}{c}{3\%} \\

Dataset & \makecell[c]{Model} &
 CD & P2M & CD & P2M & CD & P2M &
 CD & P2M & CD & P2M & CD & P2M \\
\midrule

\multirow{9}{*}{PU \cite{yu2018punet}} 
    & Bilateral \cite{fleishman2003bilateral} &
        3.646 & 1.342 & 5.007 & 2.018 & 6.998 & 3.557 &
        0.877 & 0.234 & 2.376 & 1.389 & 6.304 & 4.730
        \\
    & Jet \cite{cazals2005jetsfit}   &
        2.712 & 0.613 & 4.155 & 1.347 & 6.262 & 2.921 &
        0.851 & 0.207 & 2.432 & 1.403 & 5.788 & 4.267 
        \\
    & MRPCA \cite{mattei2017MRPCA} &
        2.972 & 0.922 & 3.728 & 1.117 & 5.009 & 1.963 &
         0.669 &  0.099 & 2.008 & 1.033 & 5.775 & 4.081 
        \\
    & GLR \cite{zeng2019GLR}   &
        2.959 & 1.052 & 3.773 & 1.306 & 4.909 & 2.114 &
        0.696 & 0.161 & 1.587 & 0.830 & 3.839 & 2.707 
        \\
\cmidrule{2-14}
    & PCN \cite{rakotosaona2020PCN}   &
        3.515 & 1.148 & 7.467 & 3.965 & 13.067 & 8.737 &
        1.049 & 0.346 & 1.447 & 0.608 & 2.289 & 1.285 
        \\
    & GPDNet\cite{pistilli2020learning}  &
        3.780 & 1.337 & 8.007 & 4.426 & 13.482 & 9.114 &
        1.913 & 1.037 & 5.021 & 3.736 & 9.705 & 7.998 
        \\
    & DMR \cite{luo2020DMR}  &
        4.482 & 1.722 & 4.982 & 2.115 & 5.892 & 2.846 &
        1.162 & 0.469 & 1.566 & 0.800 & 2.432 & 1.528 
        \\
    &  Score \cite{luo2021score} &
         2.521 &  0.463 &  3.686 & 1.074 &  4.708 &  1.942 &
         0.716 &  0.150 &  1.288 &  0.566 &  1.928 &  1.041 
        \\
\cmidrule{2-14}
    & \bf Ours &
        \bf 2.353 & \bf 0.306 & \bf 3.350 & \bf 0.734 & \bf 4.075 & \bf 1.242 &
         \bf 0.649 &  \bf 0.076 & \bf 0.997& \bf 0.296 & \bf 1.344 & \bf 0.531
        \\
    
\midrule\midrule

\multirow{9}{*}{PC \cite{rakotosaona2020PCN}} 
    & Bilateral \cite{fleishman2003bilateral} & 
        4.320 & 1.351 & 6.171 & 1.646 & 8.295 & 2.392 &
        1.172 & 0.198 & 2.478 & 0.634 & 6.077 & 2.189 
        \\
    & Jet \cite{cazals2005jetsfit}   & 
        3.032 &  0.830 & 5.298 & 1.372 & 7.650 & 2.227 &
        1.091 & 0.180 & 2.582 & 0.700 & 5.787 & 2.144 
        \\
    & MRPCA \cite{mattei2017MRPCA} & 
        3.323 & 0.931 &  4.874 & 1.178 &  6.502 & 1.676 &
        0.966 & 0.140 & 2.153 & 0.478 & 5.570 & 1.976
        \\
    & GLR \cite{zeng2019GLR}   & 
        3.399 & 0.956 & 5.274 &  1.146 & 7.249 &  1.674 &
        \bf 0.964 & \bf 0.134 & 2.015 & 0.417 & 4.488 & 1.306 
        \\
\cmidrule{2-14}
    & PCN \cite{rakotosaona2020PCN}   & 
        3.847 & 1.221 & 8.752 & 3.043 & 14.525 & 5.873 &
        1.293 & 0.289 & 1.913 & 0.505 & 3.249 & 1.076 
        \\
    & GPDNet \cite{pistilli2020learning}  & 
        5.470 & 1.973 & 10.006 & 3.650 & 15.521 & 6.353 &
        5.310 & 1.716 & 7.709 & 2.859 & 11.941 & 5.130
        \\
    & DMR \cite{luo2020DMR}  & 
        6.602 & 2.152 & 7.145 & 2.237 & 8.087 & 2.487 &
        1.566 & 0.350 & 2.009 & 0.485 & 2.993 & 0.859 
        \\
    & Score \cite{luo2021score} & 
         3.369 &  0.830 &  5.132 &  1.195 &  6.776 &  1.941 &
         1.066 &  0.177 &  1.659 &  0.354 &  2.494 &  0.657 
        \\
\cmidrule{2-14}
    & \bf Ours & 
         \bf 2.873 &  \bf 0.783 & \bf 4.757 &  \bf1.118 & \bf 6.031 & \bf 1.619 &
         1.010 &  0.146 & \bf 1.515 & \bf 0.340 & \bf 2.093 & \bf 0.573 
        \\
    
\bottomrule

\end{tabular}
}
\end{center}
\caption{{\bf Comparison among competitive denoising algorithms under isotropic Gaussian noise.} CD is multiplied by $10^4$ and P2M is multiplied by $10^4$.}
\label{table:quantitative}
\end{table*}

\section{Experiments}
\label{sec:experiments}

In this section, we evaluate the deep point set resampling model by applying
it to representative point cloud restoration tasks: point cloud denoising and
upsampling. We compare the proposed method with state-of-the-art approaches. 
\subsection{Point Cloud Denoising}
\label{subsec:experiment:denoise}

\subsubsection{Setup}
\label{subsubsec:denoise:setup}

\noindent \textbf{Datasets and Noise Models.}
Following our previous work \cite{luo2021score}, we sample 20 meshes from the training set of PUNet \cite{yu2018punet} for training. 
We use Poisson disk sampling to sample point clouds from these meshes into three resolutions: 10K, 30K and 50K. 
Then, we normalize these point clouds into the unit sphere. 
To train the denoising model, we perturb point clouds only with the commonly assumed \emph{Gaussian noise} with standard deviation from 0.5\% to 3\% of the bounding sphere's radius. 
Before feeding each point cloud into the model, we split them into patches to save GPU memory. Splitting point clouds into patches can also make our model adaptable to arbitrary resolution of point clouds. Patch size is set to be 1K during the training.

For quantitative evaluation, we choose the testing set of PUNet \cite{yu2018punet} which contains 20 different shapes, and PointCleanNet (PCN) \cite{rakotosaona2020PCN} which contains 10 shapes. 
As in the training phase, we adopt Poisson disk sampling \cite{bowers2010parallel} to sample point clouds at 10K resolution (sparse) and 50K resolution (dense) for each shape. 
During the testing phase, we feed the whole point cloud into the network for denoising instead of splitting it into patches.
Further, in order to demonstrate the generalizability of the proposed model, we perturb point clouds with various kinds of noise models for testing, including isotropic Gaussian noise, Laplace noise, discrete noise, anisotropic Gaussian noise, uni-directional Gaussian noise, uniform noise, simulated Lidar noise, as well as real-world noise. The corresponding parameters are set as follows.

\vspace{0.1in}

\noindent (1) {\bf Isotropic Gaussian noise.} We perturb point clouds with Laplace noise using the following model:
\begin{equation}
p(\vx; s) = \frac{1}{\sqrt{2\pi}s} e^{-\frac{x^2}{2s^2}}   ,
\end{equation}
The scale parameter $s$ is set to 1\%, 2\% and 3\% of the bounding sphere radius of the shape to generate point clouds at different noise levels. This setting is maintained in the following experiments, and hence will not be repeated for brevity.

\noindent (2) {\bf Laplace noise.} We perturb point clouds with Laplace noise using the following model:
\begin{equation}
p(\vx; s) = \frac{1}{2s} e^{-\frac{\left| x \right|}{s}}   ,
\end{equation}

\noindent (3) {\bf Discrete noise.} We perturb point clouds with discrete noise using the following model:
\begin{equation}
p(\vx; s) = \begin{cases}
  0.1 & \vx = (\pm s,0,0) \text{ or } (0,\pm s,0) \text{ or }  (0,0,\pm s) \\
  0.4 & \vx = (0,0,0) \\
  0 & \text{Otherwise}
\end{cases}   ,
\end{equation}
where $\vx$ denotes the coordinates of each point in the point cloud.  

\noindent (4) {\bf Anisotropic Gaussian noise.}
This is realized by setting the covariance matrix of the 3D Gaussian distribution to the following positive definite matrix:
\begin{equation}
\bm{\Sigma} = s^2 \times \begin{bmatrix}
1 & -\frac{1}{2} & -\frac{1}{4} \\
-\frac{1}{2} & 1 & -\frac{1}{4} \\
-\frac{1}{4} & -\frac{1}{4} & 1 \\
\end{bmatrix} .
\end{equation} 

\noindent (5) {\bf Uni-directional Gaussian noise.} 
This is realized by only perturbing the $x$-component of point clouds with Gaussian noise.

\noindent (6) {\bf Uniform noise.}
This is realized by employing the uniform distribution on a 3D ball to generate noise:
\begin{equation}
    p(\vx; s) = \begin{cases}
        \frac{3}{4\pi s^3} & \| \vx \|_2 \le s \\
        0 & \text{Otherwise}
    \end{cases}.
\end{equation}  

\noindent (7) {\bf Simulated Lidar noise.}
We adopt a virtual Velodyne HDL-64E2 scanner provided by the Blensor simulation package \cite{gschwandtner2011blensor} to acquire noisy point clouds. 
The noise level is set as 1\% since this experiment is for qualitative evaluation.

\noindent (8) {\bf Real-world noise.}
We also test on the \emph{Paris-rue-Madame} dataset \cite{serna2014paris} obtained from the real world using laser scanners for qualitative evaluation. 

\vspace{0.1in}
\noindent \textbf{Baselines.} We compare our method to state-of-the-art point cloud denoising algorithms, including deep-learning-based methods and optimization-based methods.

Among the deep-learning-based denoisers, we choose PCN \cite{rakotosaona2020PCN}, GPDNet \cite{pistilli2020learning}, DMRDenoise (DMR) \cite{luo2020DMR} and Score-Based Denosie (Score) \cite{luo2021score}. Optimization-based denoisers include bilateral filtering \cite{digne2017bilateral}, jet fitting \cite{cazals2005jetsfit}, MRPCA \cite{mattei2017MRPCA} and GLR \cite{zeng2019GLR}. 

\vspace{0.1in}
\noindent \textbf{Metrics.} We employ two metrics commonly adopted in previous works to perform quantitative evaluation of our model: Chamfer Distance (CD) \cite{fan2017pointsetgen} and Point-to-Mesh distance (P2M) \cite{ravi2020pytorch3d}.
Since the size of point clouds varies, we normalize the denoised results into the unit sphere before computing the metrics.

\vspace{0.1in}
\noindent \textbf{Implementation Details.} We use \textit{only one} set of hyper-parameters to train a model for \textit{all} experiments except ablation studies. 
During the training, we set the learning rate to $5e^{-4}$. 
During the resampling process, we set the step size to 0.15 and the number of steps to 50. 
Besides, the denoising results of our method doesn't require any post-processing, while previous deep-learning-based denoisers such as PCN \cite{rakotosaona2020PCN} often need to inflate their results slightly to alleviate possible shape shrinkage. 

\subsubsection{Quantitative Results}
\label{sec:experiment:quant}
We first employ isotropic Gaussian noise to test our models and baselines. 
As shown in Table~\ref{table:quantitative}, our method significantly outperforms previous deep-learning-based methods in every setting. 
Compared to optimization-based models, over model also surpasses them in most settings. 
In particular, our method achieves larger gain when the noise level is higher, further validating the superiority of the proposed model.

\begin{table}
\begin{center}
\resizebox{0.5\textwidth}{!}{
\begin{tabular}{c|c| cccccc}
\toprule
\multicolumn{2}{c|}{\# Points} & \multicolumn{6}{c}{10K}  \\

\multicolumn{2}{c|}{Noise} & 
\multicolumn{2}{c}{1\%} & \multicolumn{2}{c}{2\%} & \multicolumn{2}{c}{3\%} 
\\

  Type&Model &
 CD & P2M & CD & P2M & CD & P2M  \\
\midrule
\multirow{5}{*}{Laplace }
    &MRPCA \cite{mattei2017MRPCA} &
        2.950 & 0.724 & 4.216 &  1.428 & 7.951 & 4.441   
        \\
    &GLR \cite{zeng2019GLR}   &
        3.223 & 1.121 & 4.751 & 2.090 & 7.977 & 4.773 
        \\
    &PCN \cite{rakotosaona2020PCN}   &
        4.616 & 1.940 & 11.082 & 7.218 & 20.981 & 15.922 
        \\
    & Score \cite{luo2021score} &
         2.915 &  0.674 &  4.601 &  1.799 &  6.332 &  3.271  
        \\
\cmidrule{2-8}
    \bf &Ours &
        \bf 2.663 & \bf 0.450 & \bf 3.790 & \bf 1.067& \bf 5.110 & \bf 2.017 
        \\
\midrule
\multirow{5}{*}{Discrete }
    &MRPCA \cite{mattei2017MRPCA} &
        1.522 & 0.629 & 2.353 & 0.674 &  2.607 & 0.743    
        \\
    &GLR \cite{zeng2019GLR}   &
        1.838 & 1.014 & 2.665 & 1.047 & 2.952 & 1.116 
        \\
    &PCN \cite{rakotosaona2020PCN}   &
         1.177 & 0.307 & 2.870 & 0.871 & 4.028 & 1.674 
        \\
    & Score \cite{luo2021score} &
         1.249 &  0.251 &  2.177 & 0.416 &  2.653 &  0.653  
        \\
\cmidrule{2-8}
    \bf &Ours &
        \bf 1.021 & \bf 0.163 & \bf 1.921 & \bf 0.268 & \bf 2.274 & \bf 0.431 
        \\
\midrule
\multirow{5}{*}{Aniso}
    &MRPCA \cite{mattei2017MRPCA} &
        2.676 & 0.689 &  3.605 &  1.007 & 5.108 & 2.081    
        \\
    &GLR \cite{zeng2019GLR}   &
        2.910 & 1.048 & 3.779 & 1.332 & 4.975 & 2.195  
        \\
    &PCN \cite{rakotosaona2020PCN}   &
         3.432 & 1.129 & 7.393 & 3.940 & 12.952 & 8.654
        \\
    & Score \cite{luo2021score} &
           2.470 &  0.456 &  3.682 &  1.084 &  4.776 &  2.000  
        \\
\cmidrule{2-8}
    \bf &Ours &
        \bf 2.305 & \bf 0.308 & \bf 3.345 & \bf 0.758 & \bf 4.152 & \bf 1.350
        \\
\midrule
\multirow{5}{*}{Uni-dir}
    &MRPCA \cite{mattei2017MRPCA} &
        1.712 & 0.646 & 2.564 & 0.767 &  3.237 &  1.063    
        \\
    &GLR \cite{zeng2019GLR}   &
        2.033 & 1.026 & 2.837 & 1.139 & 3.472 & 1.434  
        \\
    &PCN \cite{rakotosaona2020PCN}   &
         1.530 & 0.432 & 3.466 & 1.360 & 5.638 & 2.914 
        \\
    & Score \cite{luo2021score} &
          1.442 & 0.279 &  2.412 &  0.543 & 3.391 &  1.108  
        \\
\cmidrule{2-8}
    \bf &Ours &
        \bf 1.256 & \bf 0.196 & \bf 2.196 & \bf 0.386 & \bf 2.862 & \bf 0.686
        \\
\midrule
\multirow{5}{*}{Uniform}
    &MRPCA \cite{mattei2017MRPCA} &
        1.555 & 0.633 & 2.754 & 0.684 & 3.229 & 0.765     
        \\
    &GLR \cite{zeng2019GLR}   &
         1.850 & 1.015 & 2.948 & 1.052 & 3.400 & 1.109  
        \\
    &PCN \cite{rakotosaona2020PCN}   &
          1.205 & 0.337 & 3.378 & 1.018 & 5.044 & 1.995
        \\
    & Score \cite{luo2021score} &
          1.277 &  0.248 &  2.467 &  0.418 &  3.079 &  0.654  
        \\
\cmidrule{2-8}
    \bf &Ours &
        \bf 1.056 & \bf 0.164 & \bf 2.348 & \bf 0.275 & \bf 2.916 & \bf 0.443
        \\
    
\bottomrule

\end{tabular}
}
\end{center}
\caption{{\bf Comparison among competitive denoising algorithms under various types of noise.} CD is multiplied by $10^4$ and P2M is multiplied by $10^4$.}
\label{table:suppl}
\end{table}

Furthermore, in order to demonstrate the generalizability of our model, we also test under various kinds of noise, including Laplace noise, discrete noise, anisotropic Gaussian noise, uni-directional Gaussian noise and uniform noise. Note that, the model being tested is \textit{exactly the same model in the previous experiment which is trained only using Gaussian noise.} 
Due to the limit of space, we compare our method with relatively stronger baselines (MRPCA \cite{mattei2017MRPCA}, GLR \cite{zeng2019GLR}, PCN \cite{rakotosaona2020PCN} and Score \cite{luo2021score}). 
Experimental results in Table~\ref{table:suppl} indicate that our method not only outperforms optimization-based methods under different noise models, but also generalizes to unseen noise models significantly better  than state-of-the-art deep-learning-based methods.

\subsubsection{Qualitative Results}
\label{sec:experiment:quali}
\begin{figure*}
\begin{center}
    \includegraphics[width=1.0\textwidth]{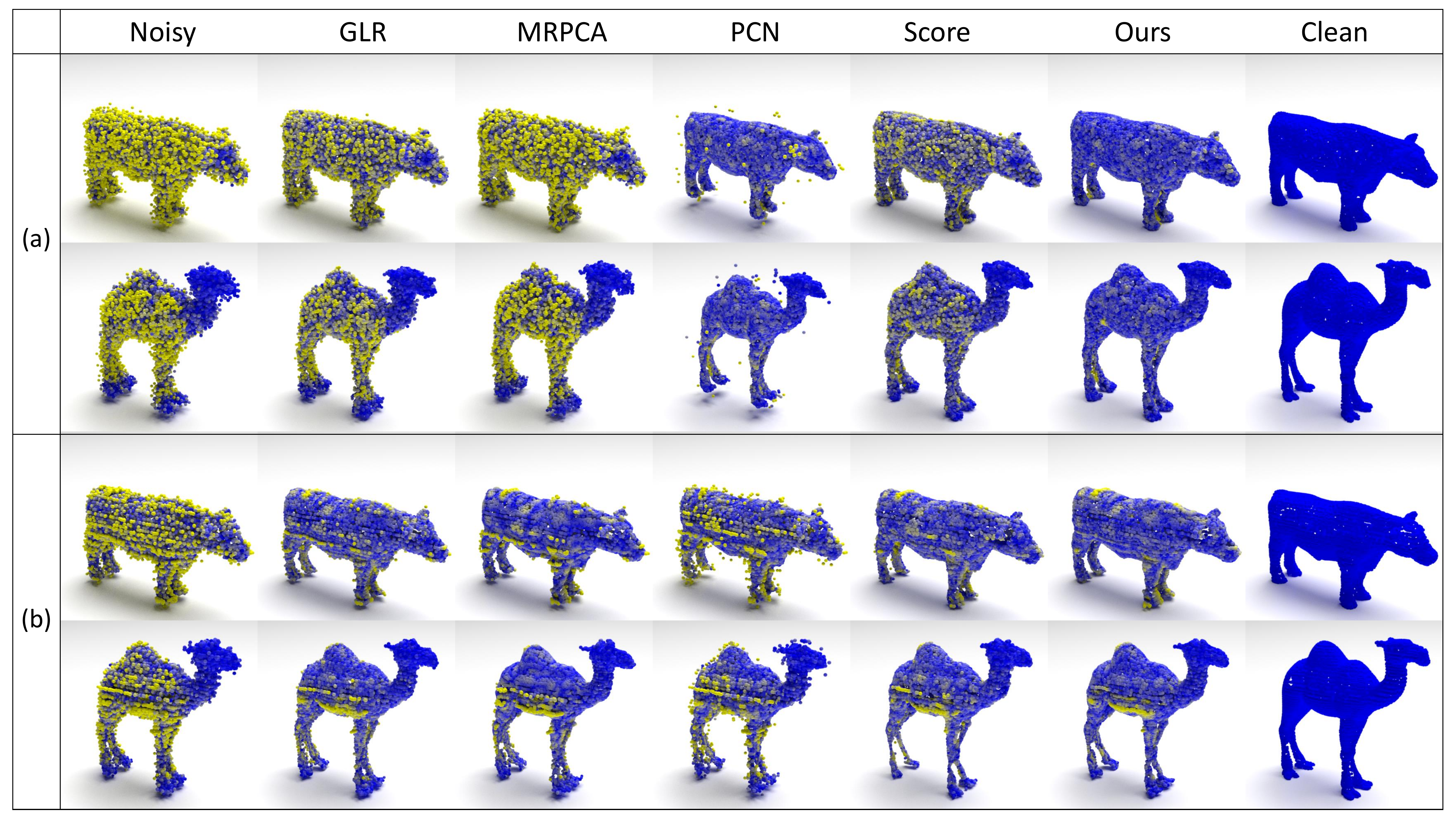}
\end{center}
\vspace{-0.2in}
\caption{{\bf Visual comparison of point cloud denoising methods under (a) Isotropic Gaussian noise, (b) simulated LiDAR noise.} Points colored yellower are farther away from the ground truth surface.}
\label{fig:visualization}
\end{figure*}

\begin{figure}
\begin{center}
    \includegraphics[width=0.5\textwidth]{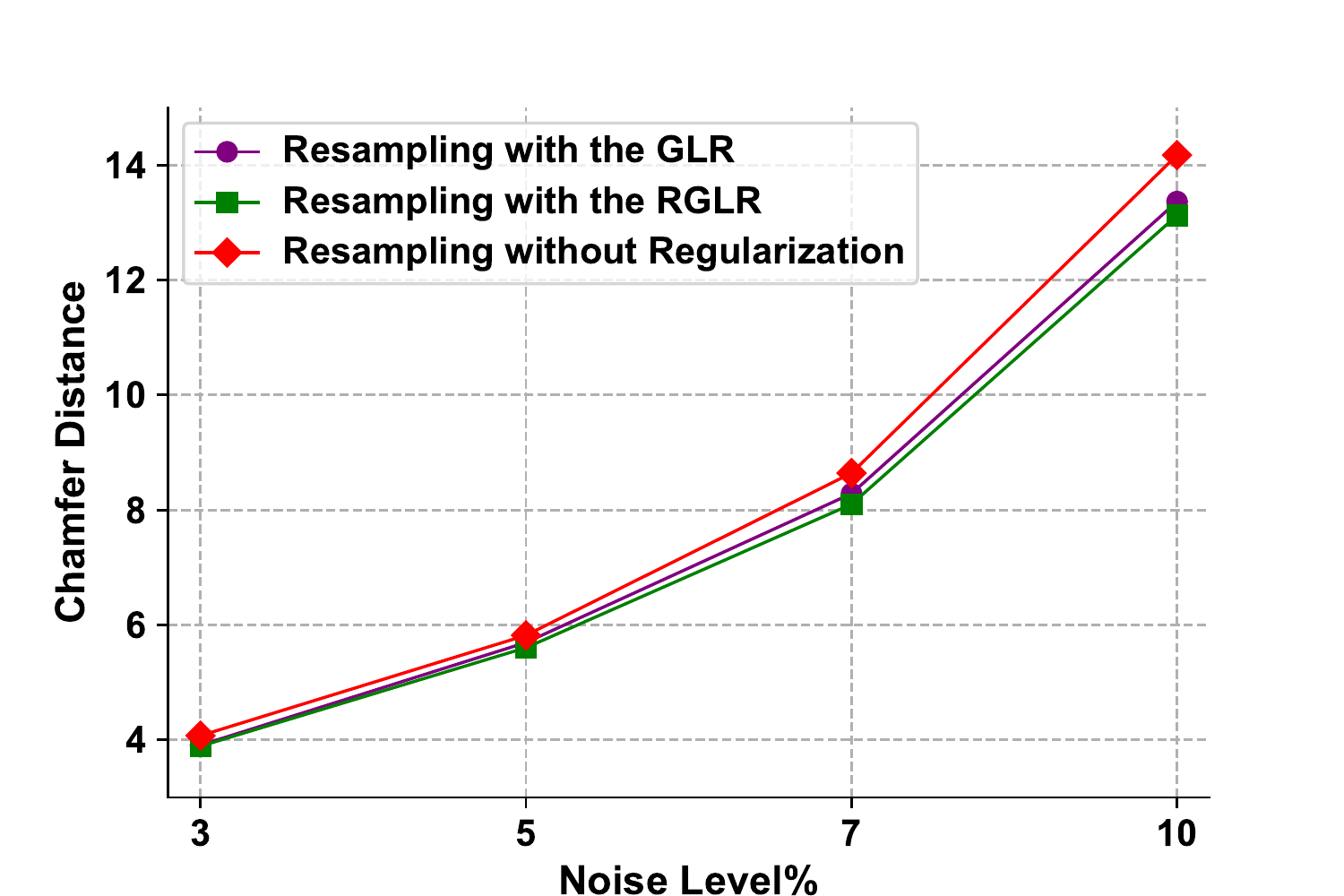}
\end{center}
\vspace{-0.2in}
   \caption{{\bf Ablation studies on resampling with regularization.} CD is multiplied by $10^4$ and P2M is multiplied by $10^4$.}
\label{fig:ablation}
\end{figure}

\begin{figure*}
\begin{center}
    \includegraphics[width=1.0\textwidth]{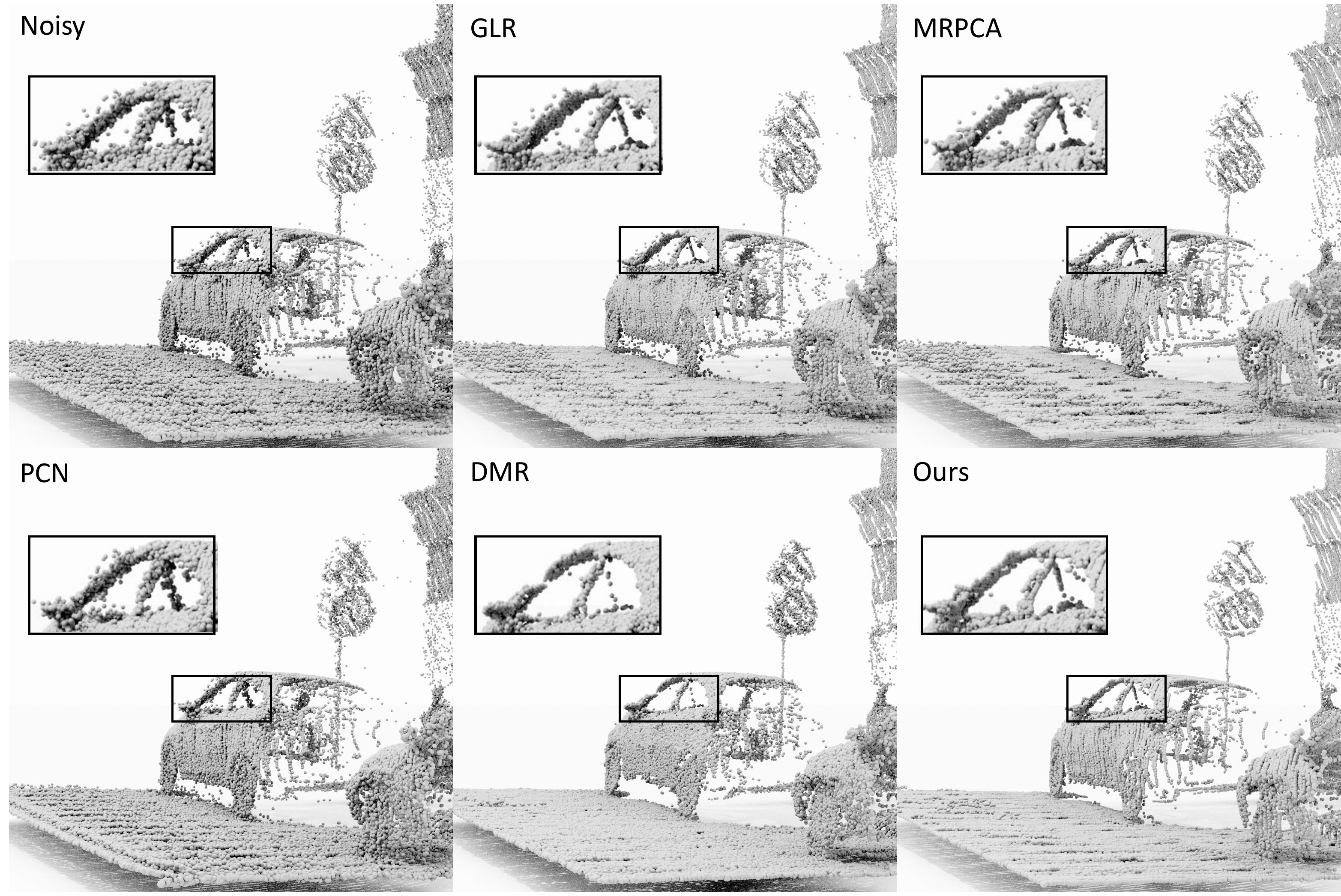}
\end{center}
\vspace{-0.2in}
\caption{{\bf Visual comparison of denoising results on the real-world dataset \textit{Paris-rue-Madame} \cite{serna2014paris}.}}
\label{fig:paris}
\end{figure*}

Figure~\ref{fig:visualization} shows the denoising results from our proposed method and competitive baselines under isotropic Gaussian noise and simulated LiDAR noise, respectively. Specifically, the level of isotropic Gaussian noise is set to 3\%, and that of the simulated Lidar noise is set to 1\%. 
The color of each point indicates its reconstruction error measured by Point-to-Mesh distance. 
Points closer to the underlying surface are colored darker, and otherwise colored brighter.

\begin{figure}
\begin{center}
    \includegraphics[width=0.5\textwidth]{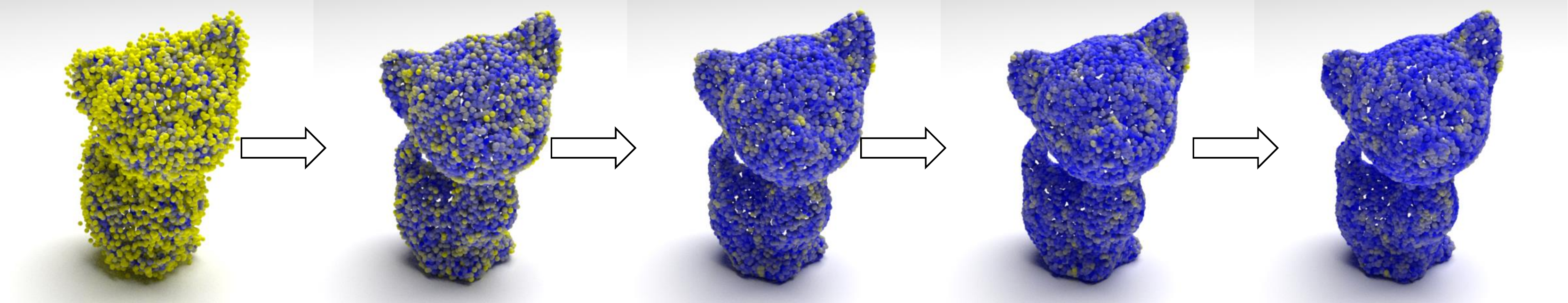}
\end{center}
\vspace{-0.2in}
   \caption{{\bf A gradient ascent trajectory of our point cloud denoising every other 10 steps.}}
\label{fig:traj}
\end{figure}
We observe from Figure~\ref{fig:visualization} that, our results are much cleaner and more visually appealing than those of other methods.
Notably, our method preserves details better than other methods and is more robust to outliers compared to other deep-learning-based methods such as PCN \cite{rakotosaona2020PCN} and Score \cite{luo2021score}.

Further, we conduct qualitative studies on the real-world dataset \emph{Paris-rue-Madame} \cite{serna2014paris}. 
Note that, since the noise-free point cloud is unavailable for real-world datasets, the error of each point cannot be computed and visualized.
As demonstrated in Figure~\ref{fig:paris}, our denoising result exhibits the most satisfactory visual quality among the competing methods. 
In particular, our result is cleaner and smoother than that of PCN \cite{rakotosaona2020PCN}, with details preserved better than DMR \cite{luo2020DMR}.

In addition, we present a denoising trajectory during the gradient ascent every other 10 steps in Figure~\ref{fig:traj}, which reveals the iterative resampling process of our method---noise reduces as points gradually converge to the mode of the distribution along the direction of the estimated gradient field. 
We see that the noise is significantly reduced during the first 10 iterations, and the denoising result almost converges at the 20th iteration, which shows the efficiency of the proposed point set resampling.


To summarize, the demonstrated qualitative results are consistent with the quantitative results in Section~\ref{sec:experiment:quant}, which again validates the effectiveness of the proposed method.


\subsubsection{Ablation Studies}
\label{sec:denoise:ablation} 

\begin{table}
\begin{center}
\resizebox{0.5\textwidth}{!}{
    \begin{tabular}{l|cccccc}
\toprule
Dataset: PU & \multicolumn{2}{c}{10K, 1\%} & \multicolumn{2}{c}{10K, 2\%} & \multicolumn{2}{c}{10K, 3\%} \\
Ablation & CD & P2M & CD & P2M & CD & P2M \\
\midrule
\bf Discon.  & 2.489 & 0.382 & 3.424 & 0.812 & 4.185 & 1.371  \\
\bf Ours &  \textbf{2.353} &  \bf 0.306 & \bf 3.350 & \bf 0.734 & \bf 4.075 & \bf 1.242 \\
\bottomrule
\end{tabular}

}
\end{center}
\caption{{\bf Ablation studies on the continuity of our model.} "Discon." represents a discontinuous version of our model without the cosine annealing employed. CD is multiplied by $10^4$ and P2M is multiplied by $10^4$.}
\label{table:ablation_con}
\end{table}
Further, we perform ablation studies to examine 1) the effectiveness of the proposed continuity model; 2) how different types of prior knowledge affect our denoising results.  

In the first ablation study, we evaluate the performance gain of the continuity model in Eq.~\ref{eq:vec_cos}. 
As shown in Table~\ref{table:ablation_con}, the continuous model outperforms the discontinuous version (\ie, removing the cosine annealing) in both metrics of CD and P2M under various noise levels, thus validating the effectiveness of the continuous model.

The second ablation study examines different prior knowledge, \ie, different regularization terms during the resampling process. As we introduced in Section~\ref{subsubsec:prior}, we focus on the Graph Laplacian Regularizer (GLR) and the Reweighted Graph Laplacian Regularizer (GLR). 
Employing the resampling algorithm with regularization in Section~\ref{subs:resample}, we compare the denoising performance without regularization, with the GLR, and with the RGLR.   
Figure~\ref{fig:ablation} demonstrates that introducing regularization into the resampling process does improve the denoising performance, and the improvement gets larger when the noise level is higher. 
This makes sense, as regularization generally plays a more important role in the challenging case of high noise levels, \ie, when data fidelity is undesirable. 
Also, we see that the RGLR performs better than the GLR, because the graph in the RGLR is learned from the updated resampled point cloud adaptively and iteratively, which is able to capture the underlying surface dynamically during the resampling process.   
Finally, we would like to emphasize that the prior knowledge is introduced in the resampling process, which requires no extra training and is thus more flexible than previous works where the regularization is considered in the training stage.


\subsection{Point Cloud Upsampling}
\label{sec:experiment:upsampling}

Given a point cloud containing $N$ points, the goal of point cloud upsampling is to infer more points and generate a dense point cloud consisting of $mN$ points, where $m$ is the upsampling ratio. 

To apply the proposed model, we first need to roughly upsample it into a denser point cloud with $mN$ points as initialization. 
Here we adopt two different initialization strategies. 
The first one is to simply perturb the sparse point cloud by slight Gaussian noise independently for $m$ times, leading to a denser but noisy point cloud with $mN$ points. 
The second approach is to adopt a lightweight neural network as a coarse generator, leveraging on the idea of Dis-PU \cite{li2021point}. 
We will introduce the implementation detail of the network in Section~\ref{sec:upsampling:setup} 

The obtained initial dense point cloud via one of the aforementioned strategies tends to be noisy and non-uniform. 
To address this problem, we treat the original sparse yet clean point cloud as the context point cloud for the training of the gradient field learning, as it contains certain information of the supporting manifold. 
Subsequently, we feed the generated initial dense point cloud into the trained model to infer the gradient field of the dense point cloud, and update the coordinates of points iteratively according to the point set resampling algorithm in Section~\ref{subs:resample} for further refinement, thus leading to a final upsampled point cloud.

\subsubsection{Setup}
\label{sec:upsampling:setup}
\noindent \textbf{Datasets}
We use the training set of PU-GAN\cite{li2019pugan} to train our model. It contains 120 meshes covering a wide variety of objects, ranging from simple objects to complex and detailed objects. 
The dataset is divided into simple, medium and complex according to the degree of fineness. 
Specifically, during the training stage, we sample point clouds from meshes with 1,024 and 2,048 points respectively as the sparse input, and sample point clouds with 4,096 and 8,192 points respectively as their corresponding ground truth dense point clouds using Poisson disk sampling \cite{bowers2010parallel}. 
Data augmentation including random rotation and scaling is also adopted to reduce over-fitting.

For quantitative evaluation, we combine the testing sets of MPU \cite{wang2019MPU} and PU-GAN \cite{li2019pugan}. 
We also adopt Poisson disk sampling \cite{bowers2010parallel} to construct the testing set, which contains 39 point clouds sampled from meshes. 
Each point cloud consists of 2,048 points. 


\vspace{0.1in}
\noindent \textbf {Baselines} We compare our method with four state-of-the-art point cloud upsampling models: EAR \cite{huang2013bilat}, PUNet \cite{yu2018punet}, MPU \cite{wang2019MPU} and PU-GAN \cite{li2019pugan}. Among them, EAR \cite{huang2013bilat} is optimization-based which requires no supervision of data, while the rest are deep-learning-based.

For fair comparison, we re-train the three deep-learning-based models with their released codes using the same dataset and same supervision data. In particular, we re-train the MPU network with two upsampling blocks because the upsampling rate of our training set is 4x, while the original MPU has 4 upsampling blocks but requires the supervision of 16x denser point clouds.

\vspace{0.1in}
\noindent \textbf{Metrics} To examine our model, we employ three commonly used metrics to quantitatively evaluate our experimental results. The adopted metrics are: Chamfer distance (CD), point-to-mesh distance (P2M) and Hausdorff distance (HD). We also normalize the input point clouds into unit sphere before feeding them into the network since the size of point clouds varies.

\vspace{0.1in}
\noindent \textbf{Implementation Details}
In the proposed point cloud upsampling method with the first noise-perturbation-based initialization strategy described in Section~\ref{sec:experiment:upsampling}, which is referred to as "Ours-Naive", we perturb each point cloud by Gaussian noise with standard deviation of 2\% for $m$ times to acquire an initial dense point cloud. 

In our upsampling method with the second generation-based initialization strategy, which we refer to as "Ours-Gen", here are the details of the coarse generator. 
We first encode the input point cloud with a feature extractor consisting of three densely connected edge convolution \cite{wang2019dynamic} layers, leading to a feature map $\mF_p$. 
Then, a feature expansion unit is employed, which simply duplicates $\mF_p$ for $m$ times and concatenates them with a regular 2D grid to obtain $\mF_e$. 
Finally, we feed $\mF_e$ into a fully-connected regression layer to generate an initial dense point cloud. 

The parameters in the training phase and resampling phase are assigned the same as those for denoising provided in Section~\ref{subsubsec:denoise:setup}.

\subsubsection{Quantitative Results}
\begin{table}
\tiny
\begin{center}
\resizebox{0.5\textwidth}{!}{
    
\begin{tabular}{c|c|ccc}
\toprule
Rate & Method  & CD & P2M & HD \\
\midrule
\multirow{5}{*}{4x}  & 
 EAR \cite{huang2013bilat} & 3.339 & 6.575 & 4.112 \\
 & PUNet \cite{yu2018punet}  & 6.692 & 14.449 & 4.773 \\ 
 & MPU \cite{wang2019MPU}   & {2.348} & 2.156 & 1.362 \\ 
 & PU-GAN \cite{li2019pugan}  & 2.489 & 2.475 & 3.898\\
 & Ours-Naive  & 2.901  & 2.598 & 1.522 \\
 & Ours-Gen  &  \bf{2.324} & \bf{1.852} & \bf{1.064} \\

 \midrule
 \multirow{5}{*}{8x}  & 
 EAR \cite{huang2013bilat}  & 2.426 & 7.427 & 4.325 \\
 & PUNet \cite{yu2018punet}  & 8.401 & 15.768 & 8.327 \\ 
 & MPU \cite{wang2019MPU}   & 1.542 & 1.828 & 1.484 \\ 
 & PU-GAN \cite{li2019pugan}   & 1.818 & 2.634 & 4.827\\
 & Ours-Naive & 1.677  & 1.577 & 1.246 \\
 & Ours-Gen &  \bf{1.532} & \bf{1.496} &\bf{1.114} \\
 \midrule
 \multirow{5}{*}{16x}  & 
 EAR \cite{huang2013bilat}  & 1.825 & 7.818 & 4.577 \\
 & PUNet \cite{yu2018punet}  & 7.274 & 12.784 & 8.710 \\ 
 & MPU \cite{wang2019MPU}  & 0.946 & 1.282 & 1.473 \\ 
 & PU-GAN \cite{li2019pugan} & 1.099 & 2.129 & 4.963\\
 & Ours-Naive & 0.980  &\bf{ 1.079} & 1.406 \\
 & Ours-Gen &  \bf{0.840} &  1.206 & \bf{1.152} \\

\bottomrule
\end{tabular}
}
\end{center}
\caption{{\bf Comparison among our model and state-of-the-art point cloud upsampling methods.} CD is multiplied by $10^4$, P2M is multiplied by $10^5$ and HD is multiplied by $10^3$. }
\vspace{-0.2in}
\label{table:ups}
\end{table}
We show the quantitative comparison in Table~\ref{table:ups}. 
Our method, especially the variant of "Ours-Gen", achieves state-of-the-art performance over all the upsampling rates. 
In particular, "Ours-Gen" achieves comparatively significant performance improvement under the metric of P2M, which indicates that our method is beneficial to the preservation of the underlying structure. 
Besides, "Ours-Naive" outperforms competitive methods at high upsampling rates (\ie, 8x, 16x). 
This is because "Ours-Naive" upsamples point clouds in a one-pass manner, while other methods upsample point clouds progressively to reach the high resolution, which may introduce error propagation.  


\subsubsection{Ablation Study}
\begin{table}
\begin{center}
\tiny
\resizebox{0.5\textwidth}{!}{
    
\begin{tabular}{c|c|ccc}
\toprule
Rate & Method  & CD & P2M & HD \\
\midrule
\multirow{2}{*}{4x}   &
  Coarse-Gen   & 2.769 & 3.262 & 1.322 \\ 
 & Ours-Gen  & \bf 2.324 & \bf 1.852 & \bf 1.064 \\
 \midrule
 \multirow{2}{*}{8x}  &
  Coarse-Gen   & 2.161 & 3.124 & 1.508 \\ 
 & Ours-Gen  & \bf 1.532 &  \bf 1.496 & \bf 1.114 \\

 \midrule
 \multirow{2}{*}{16x}  &
  Coarse-Gen   & 2.132 & 4.483 & 2.079 \\ 
 & Ours-Gen  & \bf 0.840 & \bf 1.206 & \bf 1.152 \\

\bottomrule
\end{tabular}
}
\end{center}
\caption{{\bf Ablation studies to examine the effect of the coarse generator.} }
\label{table:ablation_ups}
\end{table}

To examine the effect of the coarse generator, we also conduct ablation studies to show that only using the lightweight network of coarse generator can hardly obtain a high-quality dense point cloud especially when the upsampling rate is high. 
As presented in Table~\ref{table:ablation_ups}, our model "Ours-Gen" achieves significantly better performance than the coarse generator "Coarse-Gen", especially at high upsampling rates such as 8x and 16x. 
This is because the coarse generator is only trained with data for 4x upsampling rate.

\subsubsection{Qualitative Results}
\begin{figure}
\begin{center}
    \includegraphics[width=0.5\textwidth]{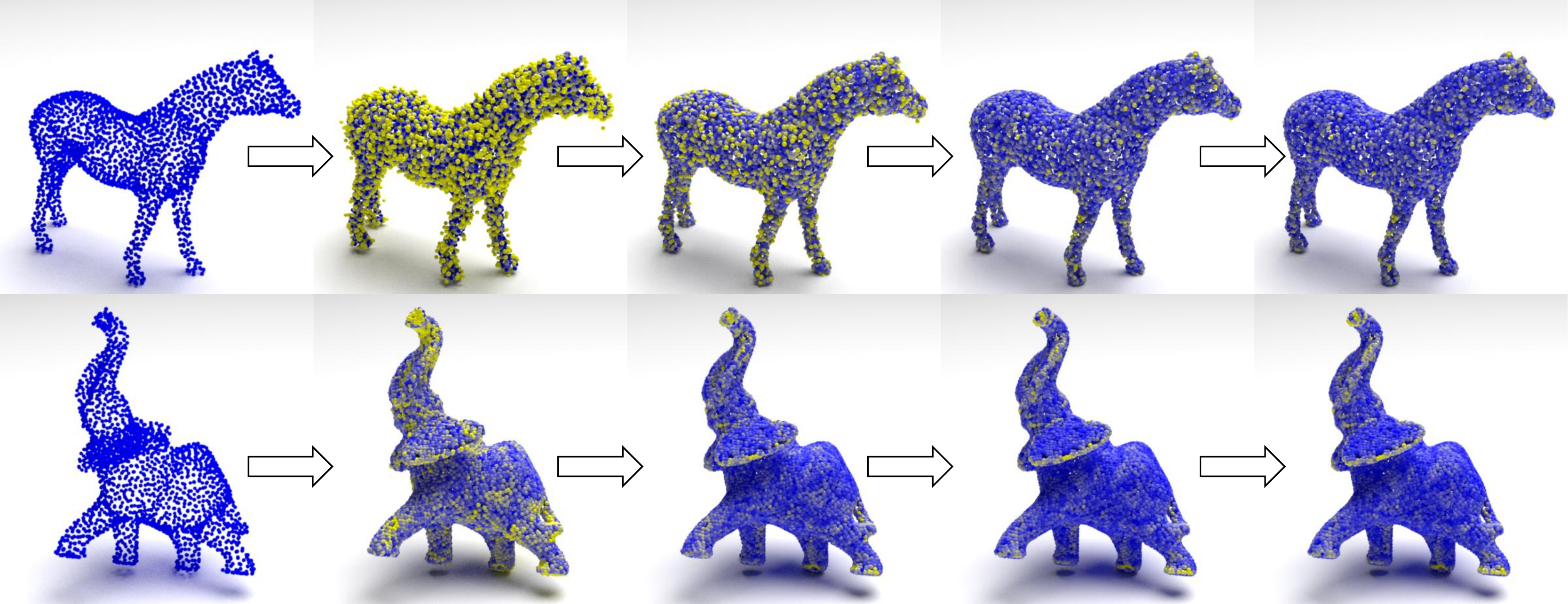}
\end{center}
\vspace{-0.2in}
   \caption{{\bf A gradient ascent trajectory of our point cloud upsampling every other 10 steps.} The first row is obtained by "Ours-Naive", while the second row is from "Ours-Gen". }
\label{fig:traj_ups}
\end{figure}

We first demonstrate the upsampling trajectory of "Ours-Naive" and "Ours-Gen" respectively in Figure~\ref{fig:traj_ups}.  
The input sparse point clouds are firstly upsampled coarsely via each variant, and then refined by the proposed resampling method based on the learned gradient field. 
We see that, the upsampling result of {\it Horse} converges after 20 steps of gradient ascent, while that of {\it Elephant} gets visually satisfactory after only 10 steps because of the better initialization.

\begin{figure*}
\begin{center}
    \includegraphics[width=1.0\textwidth]{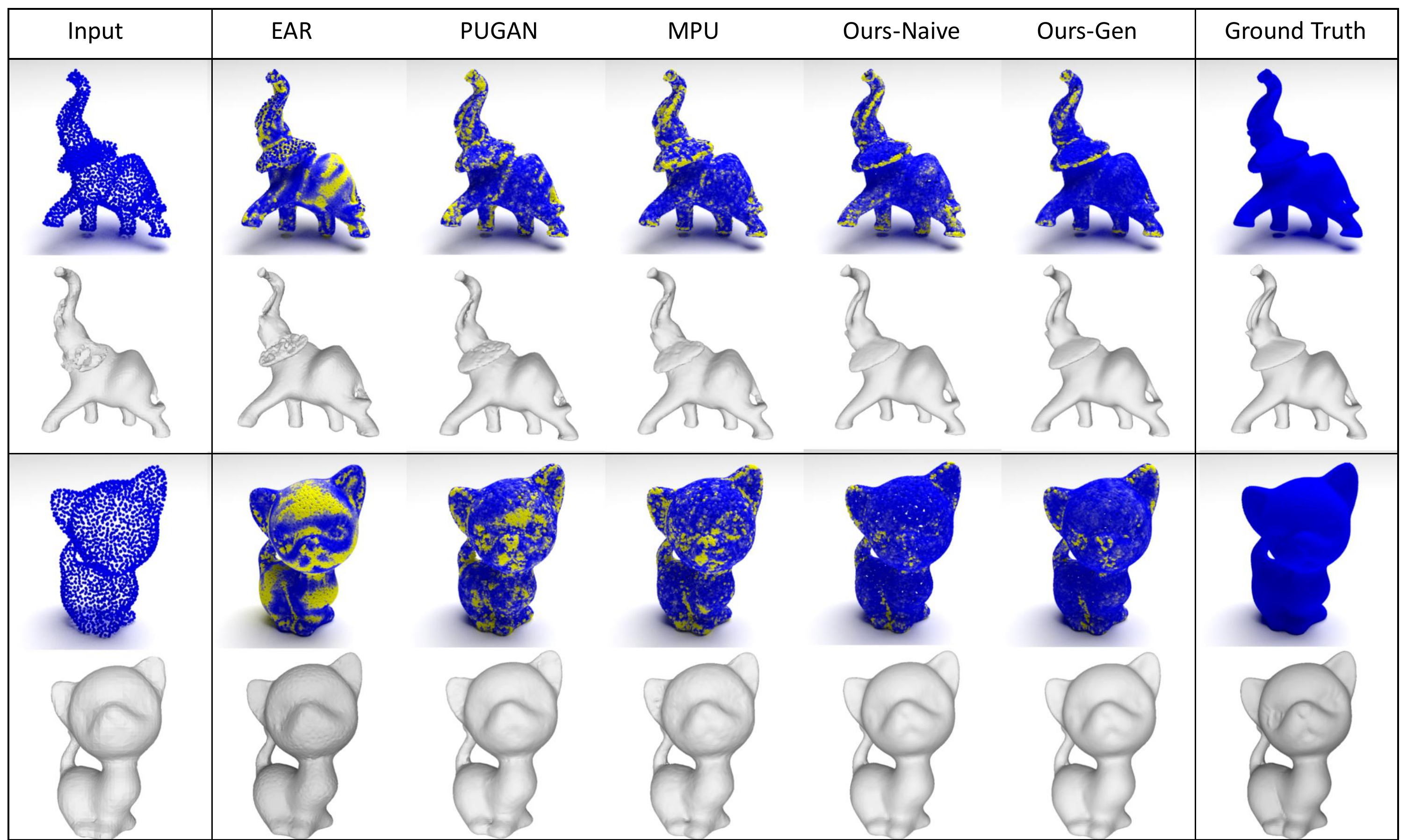}
\end{center}
\vspace{-0.2in}
\caption{{\bf Visual comparison of point cloud upsampling results from competitive methods with 8x upsampling rate and their corresponding mesh reconstructions over two shapes in the testing set.}  Points colored yellower are farther away from the ground truth surface.}
\label{fig:upsample}
\end{figure*}

Further, we visualize the results of the 8x upsampling and corresponding reconstructed meshes with screened Poisson surface reconstruction \cite{kazhdan2013screened} of two shapes "Elephant" and "Kitten" in the testing set in Figure~\ref{fig:upsample}. 
We choose competitive methods with comparatively satisfactory results for comparison. 
Figure~\ref{fig:upsample} shows that our method preserves details much better, such as around the Elephant's trunk and the Kitten's nose. 
This leads to much more satisfactory mesh reconstructions with smoother surface and more accurate details. 
Besides, our method exhibits negligible outliers.

\begin{figure}[ht]
\begin{center}
    \includegraphics[width=0.5\textwidth]{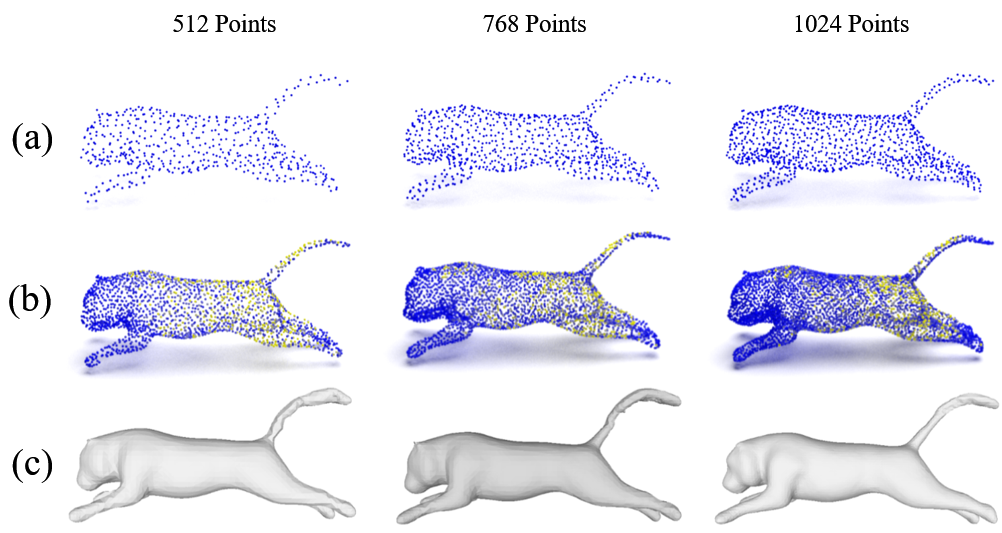}
\end{center}
\vspace{-0.2in}
\caption{{\bf Robustness of our model to the sparsity of the input point cloud.}
(a) The input with varied sparsity; (b) The corresponding upsampling results; (c) The corresponding mesh reconstructions.}
\label{fig:sparse_upsample}
\end{figure}

Moreover, we test the robustness of our model to the sparsity of input point clouds. 
We reduce the number of input points from 2048 points (the number of input points in Figure~\ref{fig:upsample}) to 1024 points, 768 points and 512 points. 
We still upsample them at the same upsampling rate
$m = 8$ and visualize the results.
Figure~\ref{fig:sparse_upsample} shows the upsampling results and corresponding mesh reconstructions. 
We see that, though the input point clouds are very sparse, our model is stable and able to preserve prominent geometric information. 
This validates the effectiveness of our model even for sparse input point clouds. 




\section{Conclusion}
\label{sec:conclusion}

In this paper, we propose a novel paradigm of point cloud resampling, which models degraded point clouds as samples from a 3D distribution and learns a global gradient field---the gradient of the log-probability density function---over the point cloud that converges points towards the underlying surface. 
We enforce the gradient field to be continuous, and introduce prior-based regularization into the resampling process to further enhance the quality of the restored point cloud. 
Based on the model, we design an efficient neural network architecture to estimate the gradient field and develop algorithms for point cloud restoration with or without regularization via gradient ascent.
Extensive experimental results validate the superiority of our model on representative tasks including point cloud denoising and upsampling.

\bibliographystyle{IEEEtran}
\bibliography{references}












\end{document}